\documentclass[journal]{IEEEtran}
\newcommand{\etal}{{\emph{et al.}}}

\usepackage[export]{adjustbox}

\usepackage{multirow}
\usepackage{booktabs}
\usepackage{makecell}
\usepackage{amssymb}
\usepackage{bbm}
\usepackage{color}

\usepackage{hyperref}
\usepackage{cite}

\usepackage{graphicx}
\usepackage{amsmath}
\usepackage{url}

\begin{document}
%
\title{Cross-Camera Distracted Driver Classification through Feature Disentanglement and Contrastive Learning}
%
%
%

\author{Luigi Celona, Simone Bianco, Paolo Napoletano\\
Department of Informatics, Systems and Communication, University of Milano-Bicocca\\viale Sarca, 336 Milano, Italy
\thanks{Corresponding author: Luigi Celona (luigi.celona@unimib.it).}
\thanks{Manuscript received April 19, 2005; revised August 26, 2015.}}

%
%

\markboth{Journal of \LaTeX\ Class Files,~Vol.~14, No.~8, August~2015}%
{Shell \MakeLowercase{\textit{et al.}}: Bare Demo of IEEEtran.cls for IEEE Journals}
%



\maketitle

\begin{abstract}
The classification of distracted drivers is pivotal for ensuring safe driving. Previous studies demonstrated the effectiveness of neural networks in automatically predicting driver distraction, fatigue, and potential hazards. However, recent research has uncovered a significant loss of accuracy in these models when applied to samples acquired under conditions that differ from the training data. In this paper, we introduce a robust model designed to withstand changes in camera position within the vehicle. Our Driver Behavior Monitoring Network (DBMNet) relies on a lightweight backbone and integrates a disentanglement module to discard camera view information from features, coupled with contrastive learning to enhance the encoding of various driver actions. Experiments conducted using a leave-one-camera-out protocol on the daytime and nighttime subsets of the 100-Driver dataset validate the effectiveness of our approach. Cross-dataset and cross-camera experiments conducted on three benchmark datasets, namely AUCDD-V1, EZZ2021 and SFD, demonstrate the superior generalization capabilities of the proposed method. Overall DBMNet achieves an improvement of 7\% in Top-1 accuracy compared to existing efficient approaches. Moreover, a quantized version of the DBMNet and all considered methods has been deployed on a Coral Dev Board board. In this deployment scenario, DBMNet outperforms alternatives, achieving the lowest average error while maintaining a compact model size, low memory footprint, fast inference time, and minimal power consumption.
\end{abstract}


%
\IEEEpeerreviewmaketitle

\section{Introduction}
\label{sec:intro}
As of 2019, road traffic accidents are the leading killer of children and youth aged 5 to 29 years and are the 12th leading cause of death when all ages are considered \cite{world2023global}. Traffic accidents pose a growing problem and are projected to rank as the seventh leading cause of global mortality by 2030~\cite{road-mortality}. 
According to statistics from the European Road Safety Observatory (ERSO), 5 to 25\% of traffic accidents in Europe stem from driver distraction~\cite{driver-distraction-summary}. 
The ERSO defines distracted driving as ``the diversion of attention away from activities critical for safe driving toward a competing activity, which may result in insufficient or no attention to activities critical for safe driving''. Commonly observed distractions include smartphone usage, eating, or conversing with passengers, all of which heighten the risk of accidents and should be rigorously avoided. Detecting these distracting behaviors through automatic in-vehicle computational systems enables early intervention to avoid distraction-related accidents. In this context, automatic Driver Monitoring Systems (DMS) have emerged as a novel Advanced Driver Assistance System (ADAS) technology to mitigate car accidents \cite{choi2016driver}. Recent legislative developments in the European Union further underline the relevance of this research direction. Specifically, Regulation (EU) 2019/2144 and the more recent Commission Delegated Regulation (EU) 2023/2590 mandate the adoption of driver drowsiness and distraction monitoring systems in all new vehicles starting from July 2024 and July 2026, respectively~\cite{eu2019,eu2023}.

DMSs often rely on models trained using data collected in controlled lab conditions, simulators, or specific vehicle setups with fixed camera placements \cite{eraqi2019driver,wang2023100,Hu_Driver2023,bianco2023platform}. However, these models frequently fail when applied to real-world scenarios where vehicle interiors, driver postures, and camera angles differ significantly from the training data. This lack of robustness leads to poor performance when the same model is deployed across different car models or camera configurations \cite{beheraLatentBodyPoseGuided2018}. In practical deployments, especially on non-production embedded platforms, DMSs must maintain reliable accuracy despite these variations—without increasing computational complexity, due to the strict resource constraints of in-vehicle hardware.


This paper addresses the challenge of cross-camera distracted driver classification by introducing the Driver Behavior Monitoring Network (DBMNet). This model is based on the use of a lightweight Convolutional Neural Network (CNN) for encoding RGB images to facilitate its deployment on commercial off-the-shelf devices. To make the model robust across different camera views, we propose a feature disentanglement module. This module separates action-relevant features from view-relevant features, allowing the proposed model to learn high quality features that are invariant to camera and view variations. To achieve this, the module is designed as a hyper-network to compute a weighted combination of the view queries and the feature vector determined by the backbone. The weights are calculated on the basis of the similarity between the test sample and the training views. Moreover, we employ a supervised contrastive learning approach with triplet loss to enhance the learning of disentangled features. While cross-entropy loss theoretically allows the model to discern similarities within categories and differences between them, it does not explicitly guarantee this outcome. In contrast, triplet loss explicitly compels the model to identify similar features for samples within the same category and distinguish features for samples from different categories. In our model, one triplet loss brings representations of different actions captured by the same camera closer together while separating them from identical actions captured by the same camera. This improves the model's ability to discriminate between different views and enhances the effectiveness of the disentanglement module. A second triplet loss brings representations of identical actions captured by different cameras closer together while separating different actions captured by the same camera. This aids in the categorization of driver actions.

The proposed work is unique because no previous approach has explicitly addressed modeling invariance in experimental configurations. We adopt a leave-one-camera-out approach, which mirrors the realistic scenario where abundant data from various cameras is available for training, and experimentation occurs under diverse conditions. We demonstrate that our novel network and learning procedure alleviates the degradation in cross-camera distracted driver classification on the very recent 100-Driver dataset \cite{wang2023100}. Experiments on the AUCDD-V1 \cite{eraqi2019driver}, EZZ2021 \cite{ezzouhri2021robust}, and SFD \cite{statefarm2020sfd} datasets demonstrate the superior cross-dataset and cross-camera performance of our approach, highlighting its generalization capability. To the best of our knowledge, this is the first work to explicitly address cross-camera generalization in distracted driver classification through a leave-one-camera-out protocol on the 100-Driver dataset—a setup that closely mirrors real-world deployment scenarios and remains largely unexplored in prior research.
Our main contributions can be summarized as follows:
\begin{itemize}
    \item A novel disentanglement module that makes features invariant to camera views, thereby enhancing driver action classification.
    \item Two triplet losses to explicitly aid in learning disentangled representations.
    \item A model that maintains low computational complexity while outperforming more advanced methods on four benchmark datasets.
    \item Feature analysis and ablation studies provide valuable insights on feature disentanglement and the impact of different design choices.
    \item An evaluation of quantized models on a Coral Dev Board, demonstrating suitability for on-edge deployment in embedded automotive systems.
\end{itemize}

The code for this work is opened on GitHub.\footnote{\url{https://github.com/CeLuigi/DBMNet}}

\section{Related works}
Large-scale image recognition networks have proven effective in recognizing distracted driving. Abouelnaga \etal~\cite{Abouelnaga_Realtime2018} employed a genetic-weighted ensemble of four CNNs, focusing on facial and hand features. Baheti \etal~\cite{Baheti_Detection2018} optimized VGG16, reducing parameters and introducing $L_2$ regularization. Koay \etal~\cite{Koay_Convolutional2021} favored CNNs over Vision Transformers and proposed the OWIPA approach with pose estimation. Shaout \etal~\cite{Shaout_Embedded2021} used SqueezeNet for real-time distraction detection. However, in all these works cross-dataset performance was overlooked.

Specially designed models like MobileVGG with 2.2M parameters \cite{Baheti_Computationally2020} advanced real-time detection. Nguyen \etal~\cite{Duy-LinhNguyen_Driver2022} developed a lightweight CNN with adaptive feature map extraction. Li \etal~\cite{Li_Driver2022} proposed OLCMNet, an 
accurate lightweight network with an Octave-Like Convolution Mixed (OLCM) block. Liu \etal~\cite{Liu_Extremely2023} used knowledge distillation and NAS, achieving high accuracy with 0.42M parameters. Mittal \etal~\cite{Mittal_CATCapsNet2023} introduced CAT-CapsNet with impressive accuracies. Despite the reported improvements, fewer parameters may limit learning capabilities, especially in cross-dataset scenarios.

To mitigate the detrimental effects of background noise on recognition accuracy, Leekha \etal~\cite{leekhaAreYouPaying2019} proposed the use of an iterative graph-based foreground segmentation algorithm to separate the driver from the background, while Yang \etal~\cite{Xing_Driver2019} employed Gaussian Mixture Models (GMM) for body extraction. 
Qin \etal~\cite{Qin_Distracted2022} introduced D-HCNN with 0.76M parameters, focusing on HOG features. Dey \etal~\cite{deyContextdrivenDetectionDistracted2021} used object detection for multiple ROIs, enhancing generalization. Behera \etal~\cite{Behera_Deep2022} proposed a contextual modeling approach that integrates posture and semantic context. They validated their approach using a self-annotated SFD test set and provided the first cross-validation results for the SFD dataset. Wang \etal~\cite{wangDataAugmentationApproach2021} applied Fast-RCNN for DOA extraction. Keypoint-based approaches, like Li \etal~\cite{liNovelSpatialTemporalGraph2019} and Bera \etal~\cite{Bera_Attend2021}, emphasized topological information. While effective, these methods demand extensive pre-processing, impacting real-time performance. Duan \etal~\cite{duan2023enhancing} proposed S-Softmax classifier and a dynamic 2D supervision matrix for improving cross-dataset performance.

Researchers simplified training strategies for enhanced generalization. Masood \etal~\cite{Masood_Detecting2020} used pretrained-weight VGG architectures for reduced training time. Li \etal~\cite{Li_Learning2022} introduced Multi-Teacher Knowledge Distillation for lightweight models. Peng \etal~\cite{Peng_TransDARC2022} employed Few-Shot Learning for feature calibration. Recently, several approaches have leveraged self-attention mechanisms in Vision Transformers (ViTs) to address driver monitoring tasks~\cite{mohammed2024driver,yang2025domain}. Mohammed \etal~\cite{mohammed2024driver} proposed a semi-supervised lightweight hybrid ViT for accurate and efficient detection of driver distraction behaviors. DGCCL \cite{yang2025domain} utilized CLIP's image encoder for feature extraction to leverage its strong generalization capabilities. To address domain shift, it introduced center loss, and further improved cross-domain performance by integrating AAMP into the classification loss.
Finally, unsupervised methods like Li \etal~\cite{Li_New2022}, focused on improved training strategies.

In order to learn more general feature representations that can later be used for classification or clustering, researchers have investigated contrastive learning exploiting contrastive and triplet loss. 
A first group of contrastive loss-based papers uses supervised contrastive loss to align representations of the same class (i.e., action) and separate those of different classes. For example, Kopuklu \etal~\cite{kopuklu2021driver} proposed a contrastive learning approach to learn a metric to differentiate normal driving from anomalous driving and introduced a new Driver Anomaly Detection (DAD) dataset with front and top drivers views in both infrared and depth modalities. 
Koay \etal~\cite{koay2023contrastive} proposed a deep contrastive learning approach exploiting multi-view and multi-modal data of DAD dataset to identify normal and anomalous driving in an open set manner, fusing together the results of different views and modalities.
\\
More often, instead, they rely on self-supervised contrastive learning to learn generalizable embeddings without explicit labels for every behavior and typically use temporal or multimodal consistency, or augmentations, to form positive and negative pairs.
Yang \etal~\cite{Yang_Quantitative2023} proposed a weakly supervised contrastive learning framework, in which distracted behaviors are identified based on their distances from the normal driving representation set.
Hu \etal~\cite{Hu_Driver2023} extended the previous approach by proposing a novel clustering supervised contrastive loss to optimize the
distribution of the extracted representation vectors, so that normal representations are better clustered while abnormal ones are separated.
Khan \etal~\cite{khan2022supervised} presented a new supervised contrastive loss function along with considerations for including projection head and refining labels to improve the detection of anomalous driving behaviors from videos.

Triplet loss-based papers instead construct triplets (anchor, positive, negative) to ensure that feature embeddings of similar driver actions (i.e., same class) are closer and those of different actions are farther apart.
Okon \etal~\cite{okon2017detecting} used the triplet loss with an offline selection of hard triplets to finetune a neural network pre-trained with classification loss.
Liu \etal~\cite{liu2021tml} used the triplet loss in a multi-task learning strategy to force the networks to explore global
information by multiple tasks. The negative sample in each triplet is generated by randomly shuffling the local regions of each given input.

Compared to previous research, this paper tackles the issue of distracted drivers by enhancing model robustness without increasing computational demands. In particular, a feature disentanglement module is introduced which is guided by two triplet losses: the first one, applied in a hyper-network fashion, encourages the removal of camera view information from the learned features; the second one enforces that similar driver actions have closer feature embeddings, while the representations of different actions are further apart. Moreover, this is the first work to conduct comprehensive experiments using the leave-one-camera-out strategy, accurately reflecting real-world scenarios where the recognition model is trained under certain conditions and tested under different ones. This rigorous setup ensures that our findings are both practical and applicable to real-world conditions.

\section{The proposed Driver Behavior Monitoring Network}
\subsection{Motivation}


Image analysis-based DMSs must operate reliably in diverse real-world conditions, where camera placement and interior configuration vary significantly across vehicles \cite{Peng_TransDARC2022,wang2023100}. This cross-camera variability introduces a critical challenge: conventional supervised deep learning methods trained on a fixed viewpoint, often employing cross-entropy loss, struggle to learn robust representations capable of generalizing across different views and, therefore, their performance degrade when they are applied to unseen views. Most existing methods overlook this issue, focusing instead on maximizing accuracy within a fixed viewpoint -- an unrealistic assumption for actual deployment.
Moreover, DMSs must run efficiently on embedded hardware, where computational and memory resources are limited. Prior approaches typically either focus on model complexity reduction or on classification accuracy improvement, but rarely address both cross-camera robustness and resource efficiency. 

In this work, we address both challenges jointly. We propose a learning strategy that explicitly disentangles driver action features from viewpoint-dependent cues, enabling the model to generalize across camera perspectives while maintaining a lightweight structure suitable for real-time applications. 
We introduce the feature disentanglement module, which is based on two complementary triplet losses to guide the network to learn robust, semantically meaningful representations. The feature disentanglement module uses the first triplet loss to train an action-independent view classifier and exploits it as a hyper-network to remove of camera view information from learned features. The second triplet loss is used to further reinforce viewpoint consistency across actions. 

Our goal is not only to improve classification accuracy, but to ensure transferability and reliability in real deployment scenarios.

\subsection{Overview}
\begin{figure*}
    \centering
    \includegraphics[width=.9\linewidth]{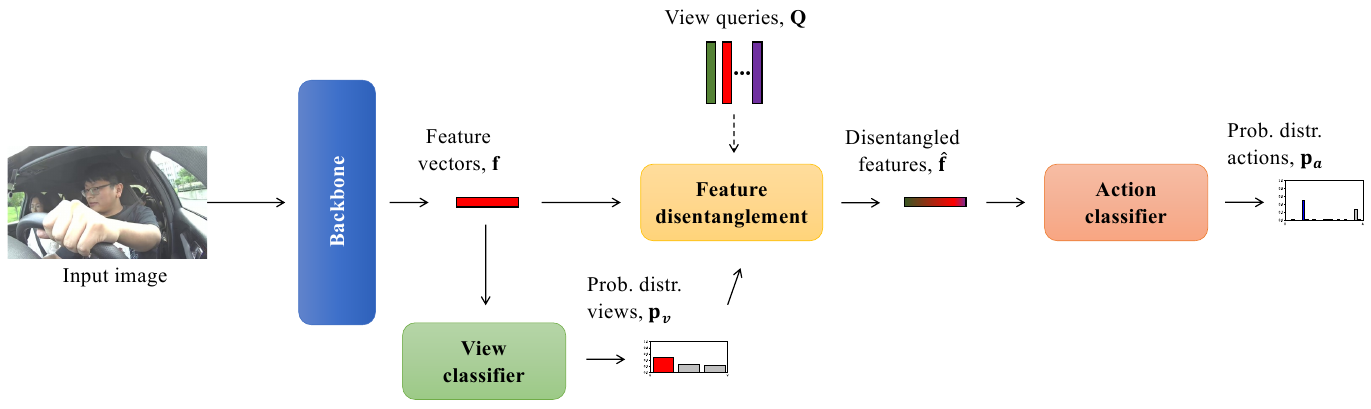}\\
    (a) \vspace{1em}\\
    \includegraphics[width=.92\linewidth]{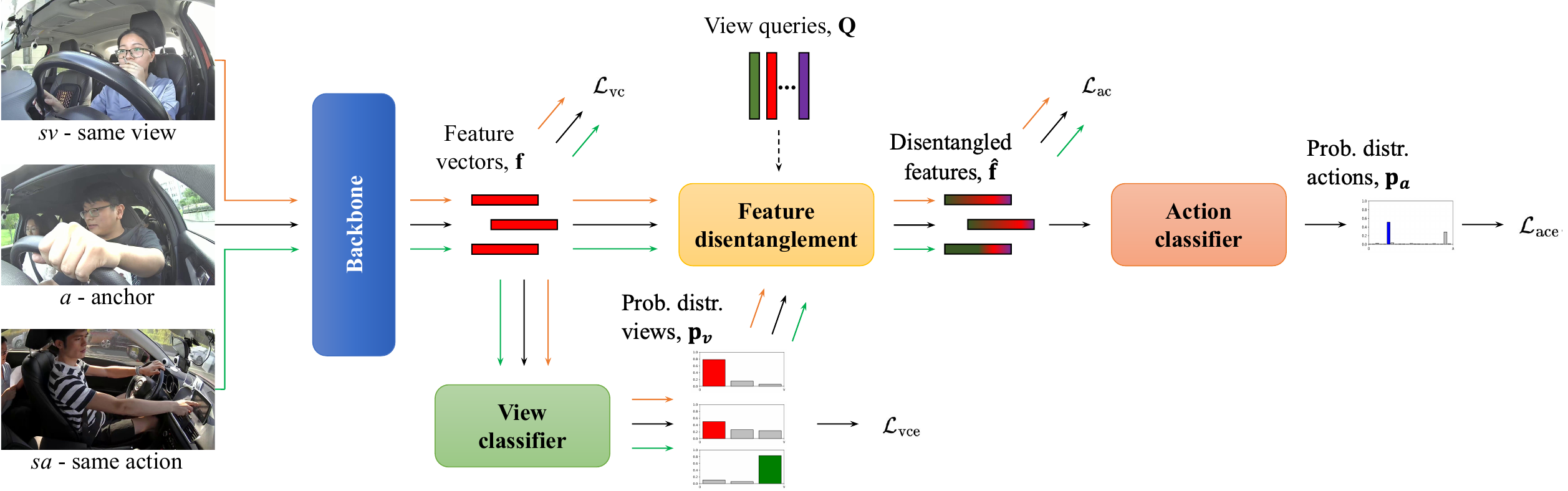} \\(b)
    \caption{The \textit{DBMNet} pipeline. At inference time (a), the backbone encodes an RGB image into a feature vector $\mathbf{f}$. The feature disentanglement module then refines these features to discard view-related information while retaining action-related details using view queries, $\textbf{Q}$, together with the probability distribution over the views. The resulting features $\mathbf{\hat{f}}$ are mapped into a driver action by the action classifier. During training (b), alongside the cross-entropy loss for action and view prediction (respectively, $\mathcal{L}_{\rm{ace}}$ and $\mathcal{L}_{\rm{vce}}$), we employ the triplet loss ($\mathcal{L}_{\rm{ac}}$ and $\mathcal{L}_{\rm{vc}}$). This triplet loss helps to learn distinct action and view representations by processing three input images: an anchor image $a$, an image $sv$ depicting the same view but different action as $a$, and an image $sa$ with the same action but from a different view as $a$.}
    \label{fig:pipeline}
\end{figure*}

Given the factors mentioned above, we present the Driver Behavior Monitoring Network (DBMNet), illustrated in Figure \ref{fig:pipeline}. This method encodes an RGB image by leveraging the lightweight CNN architecture detailed in \cite{wang2023100} as its backbone. The proposed feature disentanglement module aims to prioritize action encoding over camera-view information. This is achieved by linearly combining view queries with features estimated by the backbone. The resulting disentangled features are then forwarded to a linear layer for action classification. The model outputs two probability distributions: one for the driver's action categories and another for the viewpoint. The viewpoint distribution acts as a hyper-network for the feature disentanglement module. The model is optimized through cross-entropy losses combined with triplet losses to assist in the learning of view-invariant action features.
\subsection{The proposed model}
\paragraph{Backbone} Our model consists of a GhostNet-v1.0 \cite{han2020ghostnet} as the backbone. 
Given an input image $\mathbf{X} \in \mathbb{R}^{C\times H\times W}$, the backbone $f_b$ outputs a feature vector $\mathbf{f} = f_b(\mathbf{X}) \in \mathbb{R}^{1\times D}$ as output, where $D$ is the number of features.

\paragraph{Feature disentanglement} The output of the backbone is passed to the feature disentanglement module. This module consists of $V$ learnable view queries $\mathbf{Q} \in \mathbb{R}^{D \times V}$, where $V$ represents the number of views for the training samples. Through our learning process, we determine, for each view, which features are most relevant for characterizing the driver actions. The disentangled feature vector is obtained as a weighted combination of the view queries with the feature vector $\mathbf{f}$, determined by the backbone. The weighting, denoted as $\mathbf{w}$, is computed based on which view or views the test sample most closely resembles. In practice, estimating the disentangled features is carried out as follows:
\begin{align}
    \hat{\mathbf{f}} = \mathbf{w} \times \mathbf{f}, \\
    \mathrm{with} \quad \mathbf{w} = \mathbf{p}_v \cdot \mathbf{Q},
\end{align}
where $\hat{\mathbf{f}}$ is the disentangled feature vector used for action classification, $\mathbf{p}_v$ is the probability distribution over $V$ views estimated by the view classifier, which therefore acts as a hyper-network \cite{ha2017hypernetworks,chauhan2024brief}. The inner structure of this module might resemble memory banks, which are typically a learnable, or fixed, set for feature vectors or embeddings that the network can query and update: the idea of memory banks is to provide a module that can store feature representations so the model can retrieve and reuse them. Our feature disentanglement module instead learns a set of view queries that are independently multiplied for the feature vector coming from the backbone and linearly combined using as weights the probability distribution over the views that is estimated by the view classifier. As such, it also shares some similarities with the attention mechanism \cite{schug2024attention}.

\paragraph{Classifiers} The model consists of two classifiers, namely the Action classifier for categorizing the action performed by the driver and the View classifier for the estimation of the camera viewpoint. The Action classifier takes as input the disentangled feature vector $\hat{\mathbf{f}}$ and outputs the probability distribution over the $A$ action classes, $\mathbf{p}_a \in \mathbb{R}^{1\times A}$, as follows:
\begin{equation}
    \mathbf{p}_a = (\mathbf{W}_a \hat{\mathbf{f}}) + \mathbf{b}_a,
\end{equation}
where $\mathbf{p}_a$ denotes the probability distribution over the $A$ driver action classes, $\mathbf{W}_a \in \mathbb{R}^{D\times A}$ and $\mathbf{b}_a \in \mathbb{R}^{1\times A}$ are the weights and bias of the fully-connected head for action classification.

Instead, the View classifier processes the feature vector $\mathbf{f}$ produced directly from the backbone to produce the probability distribution, $\mathbf{p}_v$, on the $V$ views as follows: 
\begin{equation}
    \mathbf{p}_v = (\mathbf{W}_v \mathbf{f}) + \mathbf{b}_v.
\end{equation}
In the previous equations, $\mathbf{W}_v \in \mathbb{R}^{D\times V}$ and $\mathbf{b}_v \in \mathbb{R}^{1\times V}$ are the weights and bias of the fully-connected layer for view classification.

\subsection{Training procedure}
\paragraph{Training data input} During training, our model processes batches of image triplets with an approach similar to that of Siddiqui \etal~\cite{siddiqui2024dvanet}. Each triplet comprises an anchor image, denoted as $a$, which portrays a driver's action from a specific viewpoint. Another image in the triplet, referred to as $sv$ (same view), displays a different action captured from the identical viewpoint as $a$. Finally, we include an image of the same action as $a$, but captured from a distinct viewpoint, denoted as $sa$ (same action). We construct these triplets to provide both positive and negative samples for our supervised contrastive learning, which will be presented in the following section. By supplying both positive and negative samples, the model can develop a comprehensive understanding of the discriminative characteristics within each action or viewpoint category through sample comparisons. This methodology diverges from the learning-by-classification approach, wherein the loss function solely penalizes classification errors without explicitly promoting intraclass compactness and interclass separation \cite{schroff2015facenet}.

\paragraph{Loss function} Our DBMNet is trained through a composite loss function that is defined as follows:
\begin{equation}
\label{eq:loss}
    \mathcal{L} = \mathcal{L}_{\rm{ace}} + \mathcal{L}_{\rm{vce}} + \lambda_{\rm{ac}} \mathcal{L}_{\rm{ac}} + \lambda_{\rm{vc}}\mathcal{L}_{\rm{vc}}.
\end{equation}
In the previous loss, there are two cross-entropy losses mainly dedicated to the optimization of linear layers for view ($\mathcal{L}_{\rm{vce}}$) and action ($\mathcal{L}_{\rm{ace}}$) categorization. Cross-entropy losses are only applied  to the anchor image in this way:
\begin{align}
    \mathcal{L}_{\rm{vce}} = -\sum_{m=1}^Vy^{(m)}_v\mathrm{log}(p_v^{(m)}),\\
    \mathcal{L}_{\rm{ace}} = -\sum_{m=1}^Ay^{(m)}_a\mathrm{log}(p_a^{(m)}).
\end{align}
with $A$ and $V$ being the number of actions and the number of views, respectively. $p^{(m)}_a$ is the probability of the $m$-th category with ground-truth $y_a^{(m)}$ for the action of the anchor sample, while $p^{(m)}_v$ is the probability of the $m$-th category with ground-truth $y_v^{(m)}$ for the anchor sample viewpoint.

The previous two cross-entropy losses are complemented by two opposing triplet losses, whose contributions are controlled by the hyperparameters $\lambda_{\rm{ac}}$ and $\lambda_{\rm{vc}}$. Both losses rely on the triplet structure previously introduced. The difference lies in the role assumed by $sv$ and $sa$ in each loss.

In the action triplet loss (Eq. \ref{eq:loss-ac}) the objective is to learn features for action encoding that are invariant regardless of the viewpoint. Hence, $sa$ serves as the positive sample, whereas $sv$ serves as the negative sample. This choice has the purpose of bringing closer the representations of $sa$ and $a$, which have the same action. Conversely, the model maximizes the distance between $a$ and $sv$, which represent different actions from the same viewpoint:
\begin{equation}
\label{eq:loss-ac}
    \mathcal{L}_{\rm{ac}} = \sum_{n=1}^{N} [\delta + \mathcal{D}(\mathbf{\hat{f}}_a, \mathbf{\hat{f}}_{sa})-\mathcal{D}(\mathbf{\hat{f}}_a, \mathbf{\hat{f}}_{sv})],
\end{equation}
where $\hat{\mathbf{f}}_a$, $\hat{\mathbf{f}}_{sa}$, and $\hat{\mathbf{f}}_{sv}$ denote the disentangled feature outputs from the $a$, $sa$, and $sv$ images, respectively. $\mathcal{D}(\cdot, \cdot)$ is the pairwise Euclidean distance function, and $\delta$ is the margin parameter enforcing separation between positive and negative pairs. It is expected that the learned disentangled features exhibit view-invariance, as they are encouraged to be as similar as possible when derived from two different images depicting the same action regardless of the viewpoint from which they were acquired.

For the view triplet loss, the positive and negative samples are reversed:
 \begin{equation}
     \mathcal{L}_{vc} = \sum_{n=1}^{N} [\delta + \mathcal{D}(\mathbf{f}_a, \mathbf{f}_{sv})-\mathcal{D}(\mathbf{f}_a, \mathbf{f}_{sa})].
 \end{equation}

where $\mathbf{f}_a$, $\mathbf{f}_{sv}$, and $\mathbf{f}_{sa}$ are the backbone feature vectors for the anchor, same-view, and same-action samples, respectively.

In this way, the feature disentanglement module, particularly the view queries, learns to generate similar features for images acquired from the same viewpoint, regardless of the action being performed. Since our primary goal is to enhance multi-view driver action recognition, training our model to distinguish action-relevant features from view-relevant features in this manner provides cleaner and more accurate action representations that are robust to changes in the camera viewpoint.

\section{Experiments}
In this section, we present the datasets used in the experiments, describe the experimental setup, and discuss the results of our extensive experimentation.

\subsection{Dataset}
We conduct extensive experiments on 100-Driver \cite{wang2023100}. The 100-Driver dataset comprises 470,000 images illustrating 100 drivers engaged in 22 distinct actions during driving sessions. These actions include 21 distracted behaviors and one instance of normal driving (see Table \ref{tab:action-list} for the list of all actions).

\begin{table*}
    \centering
    \caption{The list of driver actions in the 100-Driver dataset \cite{wang2023100}.}
    \label{tab:action-list}
    \begin{tabular}{lp{3cm}|lp{3cm}|lp{3cm}|lp{3.5cm}}
    \toprule
    No. & Action & No. & Action & No. & Action & No. & Action \\ \midrule
        0 & Normal driving & 6 & Texting (right) & 12 & Smoking (left) & 18 & Operating GPS \\ 
        1 & Sleeping & 7 & Hair / makeup & 13 & Smoking (right) & 19 & Reaching behind\\
        2 & Yawning & 8 & Looking left & 14 & Smoking (mouth) & 20 & Hands off the steering wheel \\ 
        3 & Talk with cellphone (left) & 9 & Looking right & 15 & Drinking / Eating (left) & 21 & Talking to passenger\\ 
        4 & Talk with cellphone (right) & 10 & Looking up & 16 & Drinking / Eating (right)\\ 
        5 & Texting (left) & 11 & Looking down & 17 & Adjusting radio \\ \bottomrule
    \end{tabular}
\end{table*}

\begin{figure*}
    \centering
    \begin{tabular}{cccc}
         \includegraphics[width=.22\textwidth]{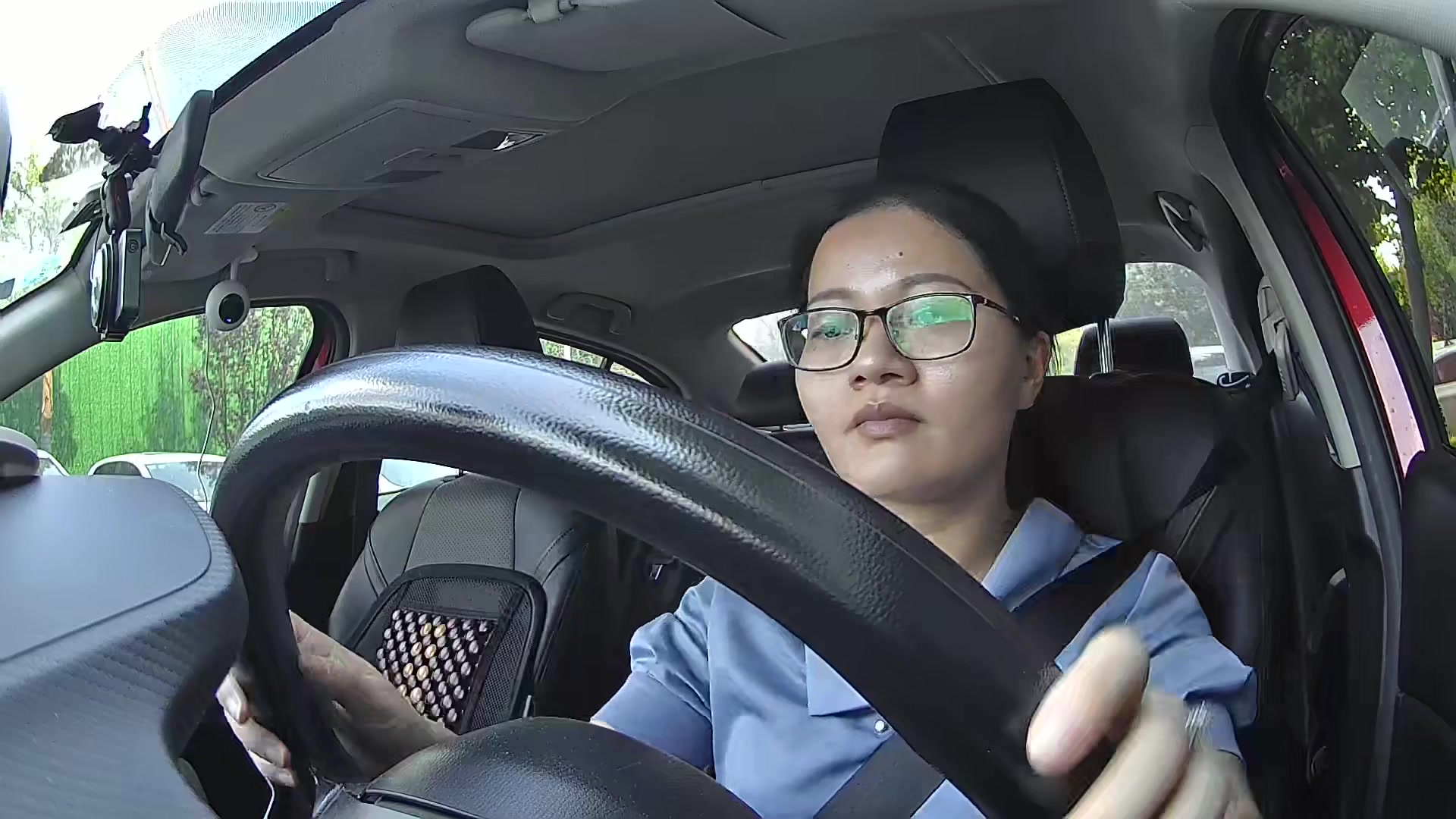} & \includegraphics[width=.22\textwidth]{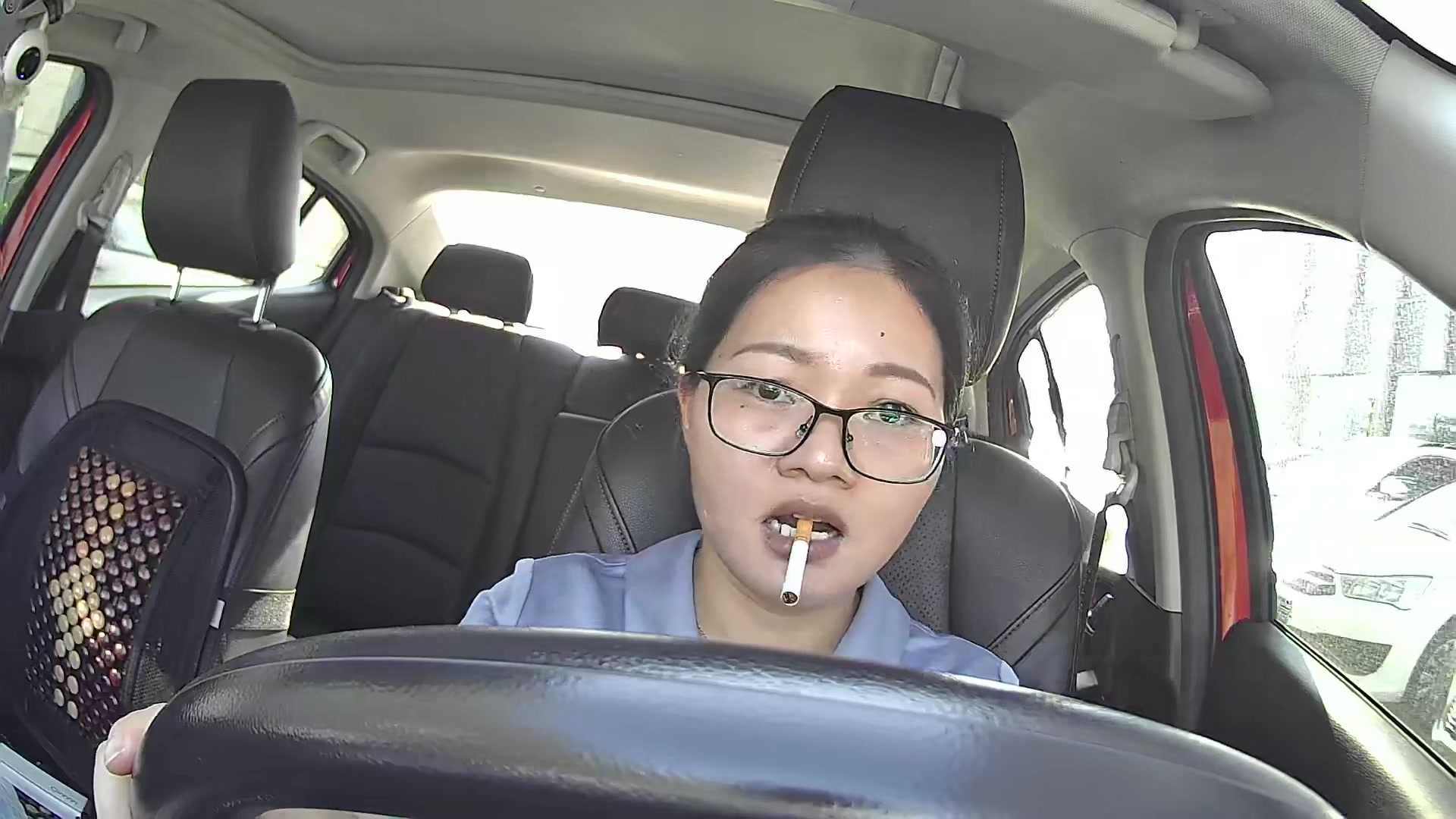} & \includegraphics[width=.22\textwidth]{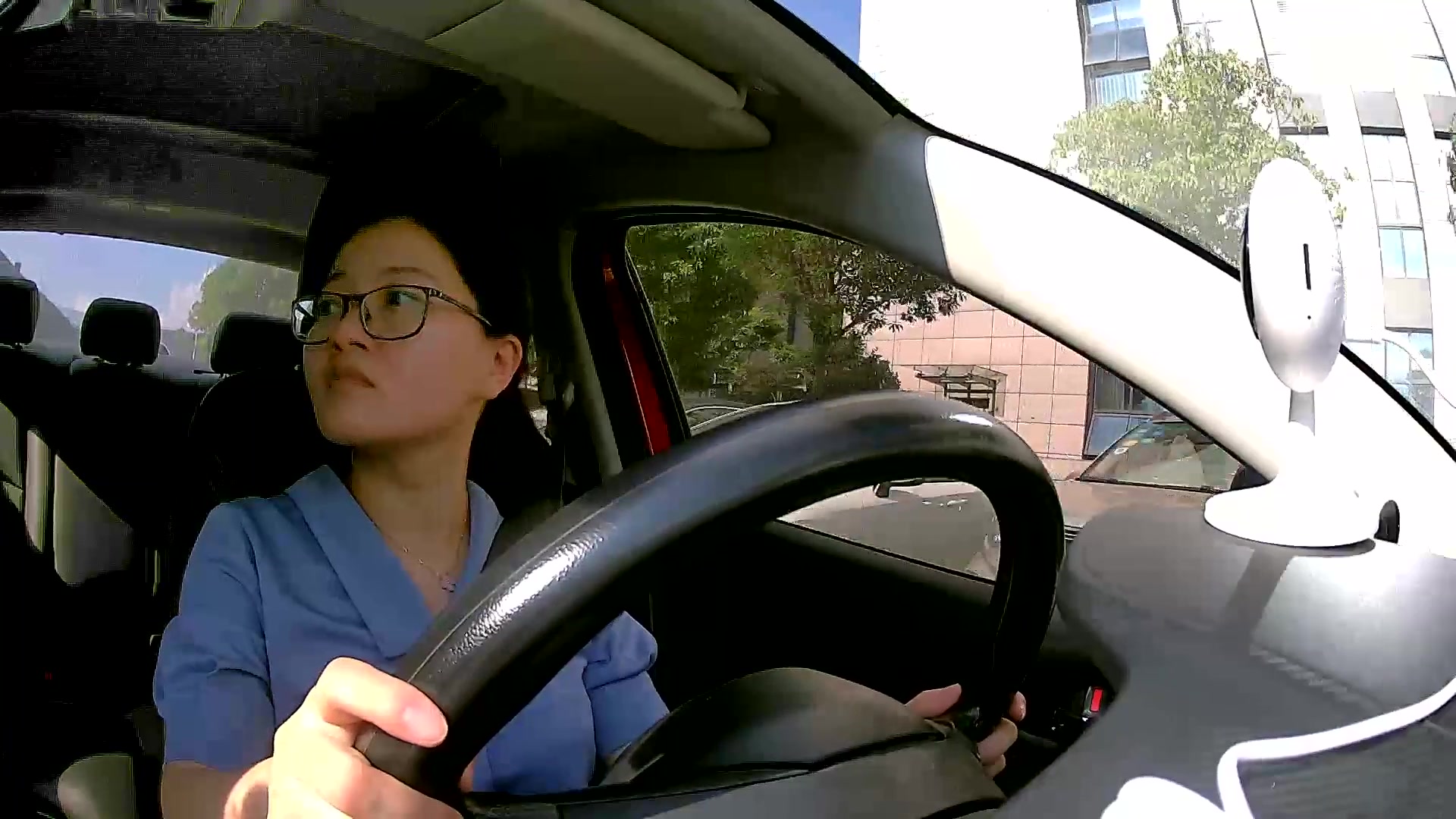} & \includegraphics[width=.22\textwidth]{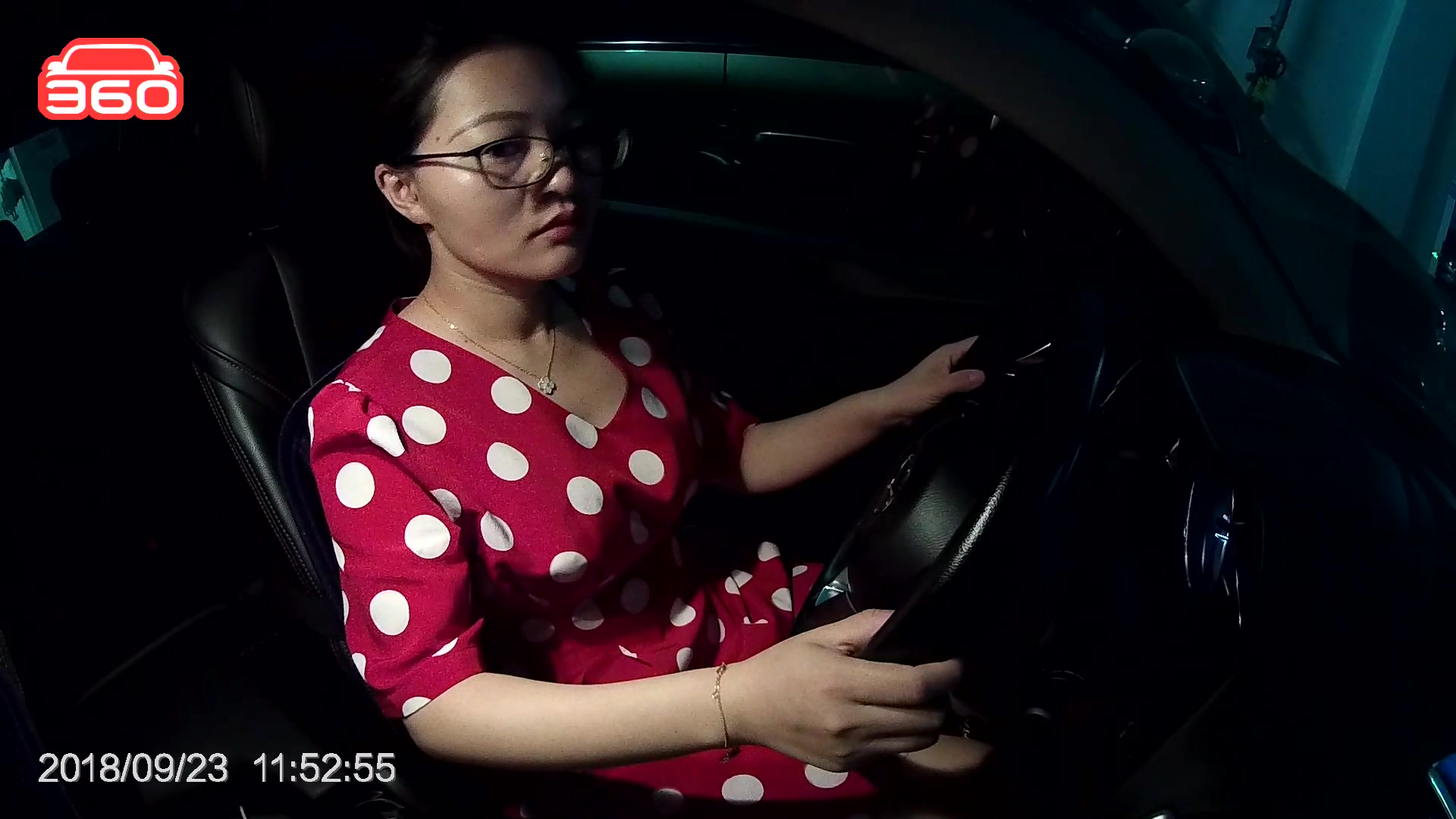} \\
         D1 & D2 & D3 & D4 \vspace{1em}\\
        \includegraphics[width=.22\textwidth]{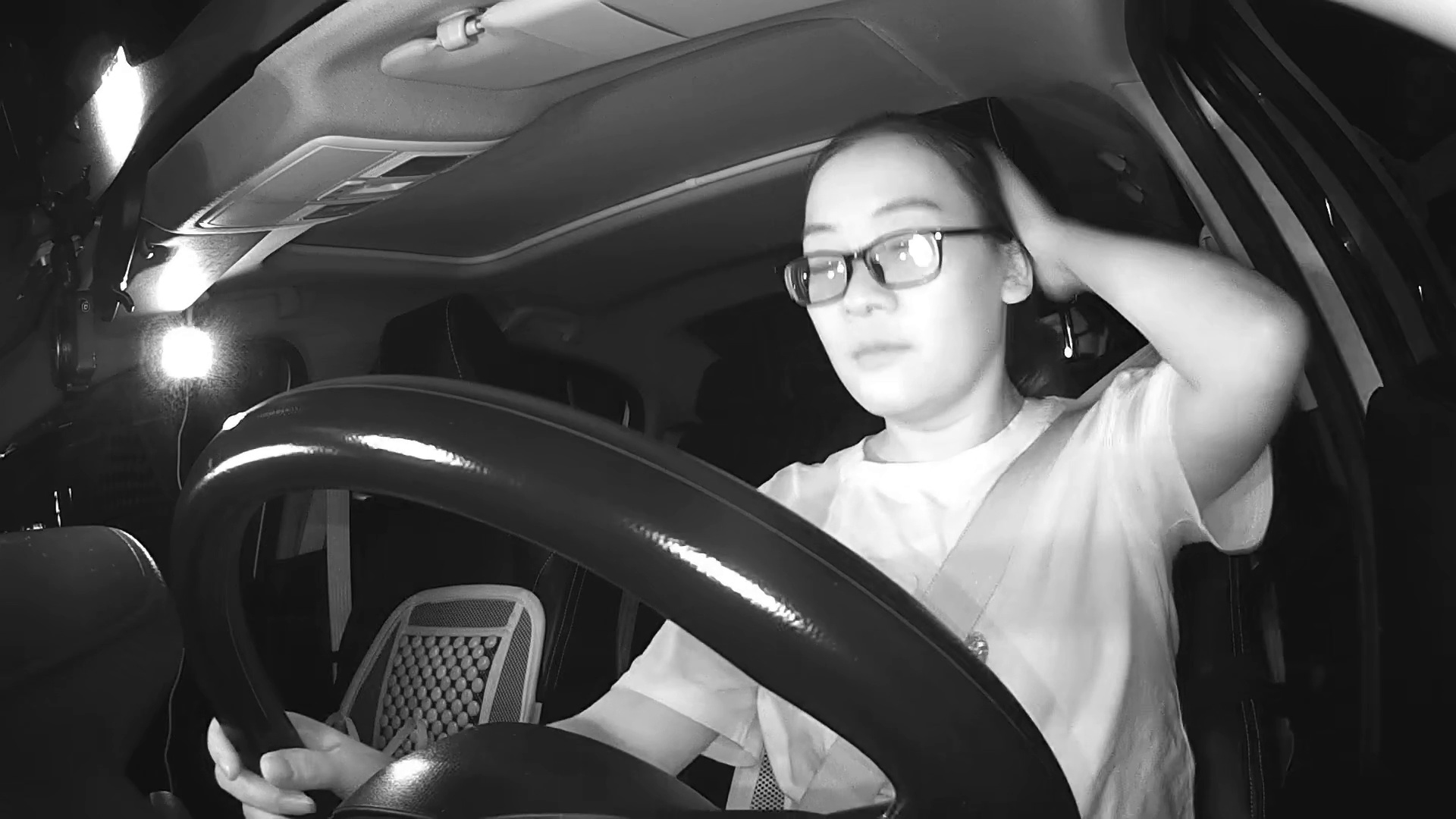} & \includegraphics[width=.22\textwidth]{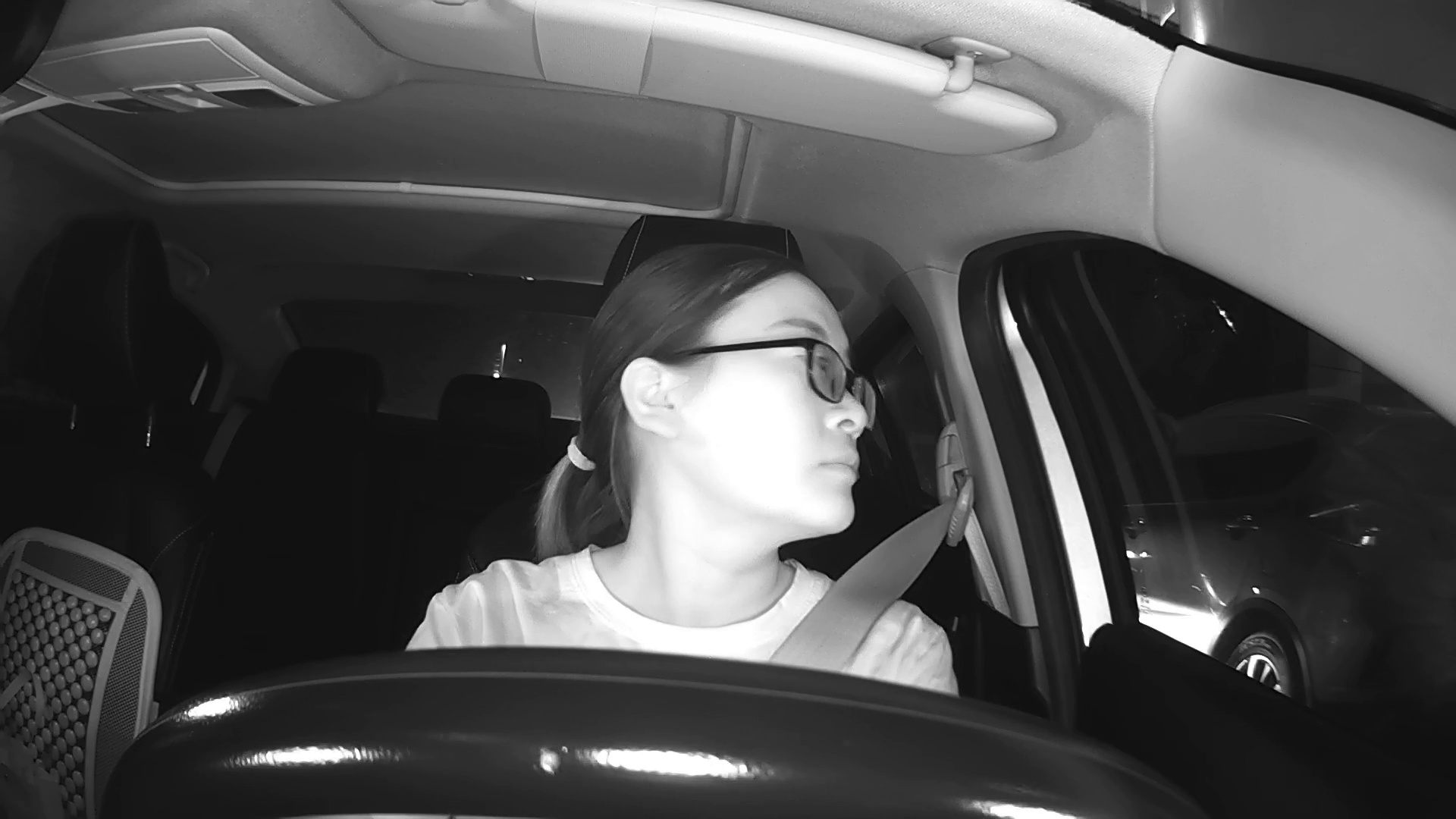} & \includegraphics[width=.22\textwidth]{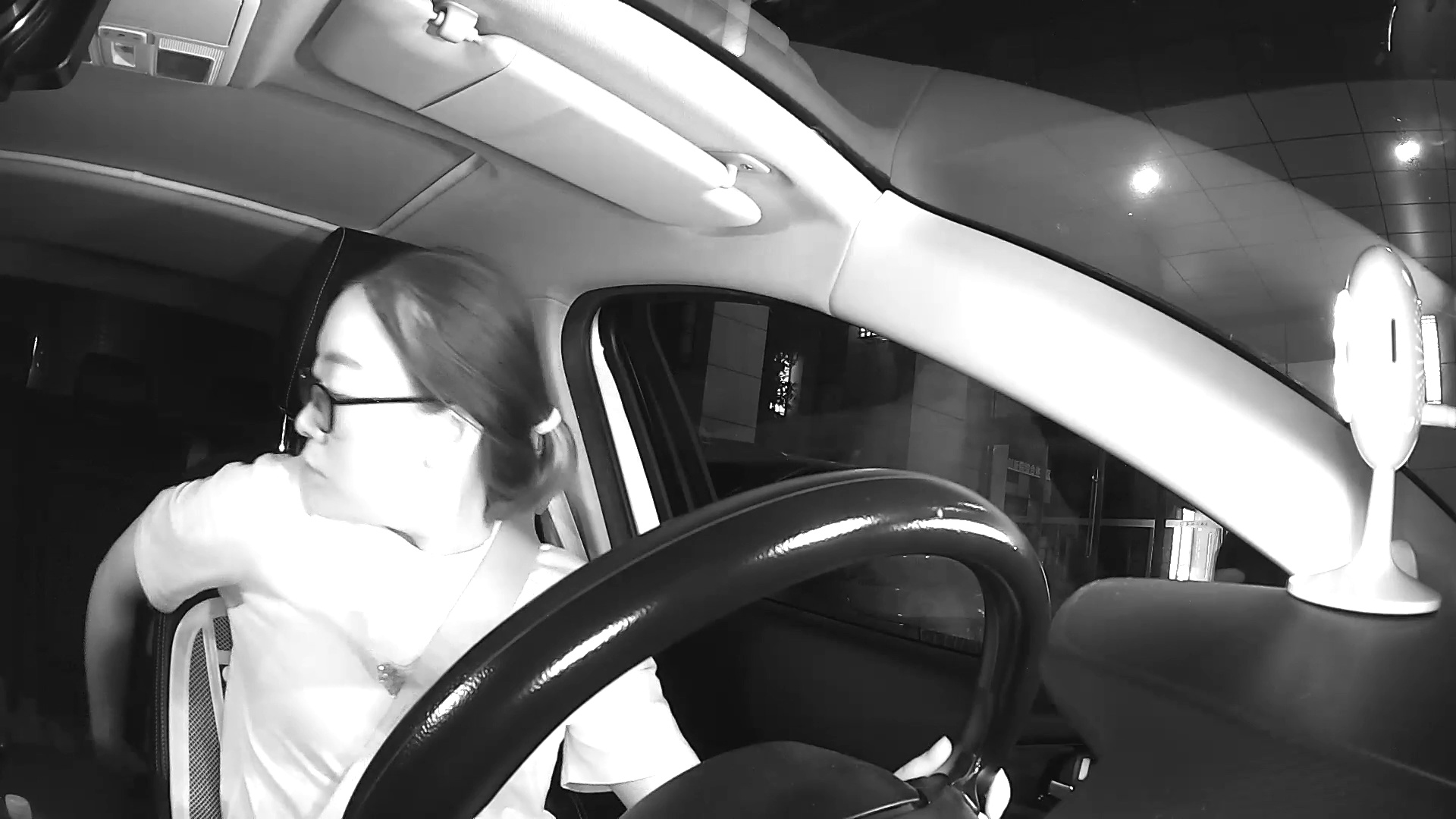} & \includegraphics[width=.22\textwidth]{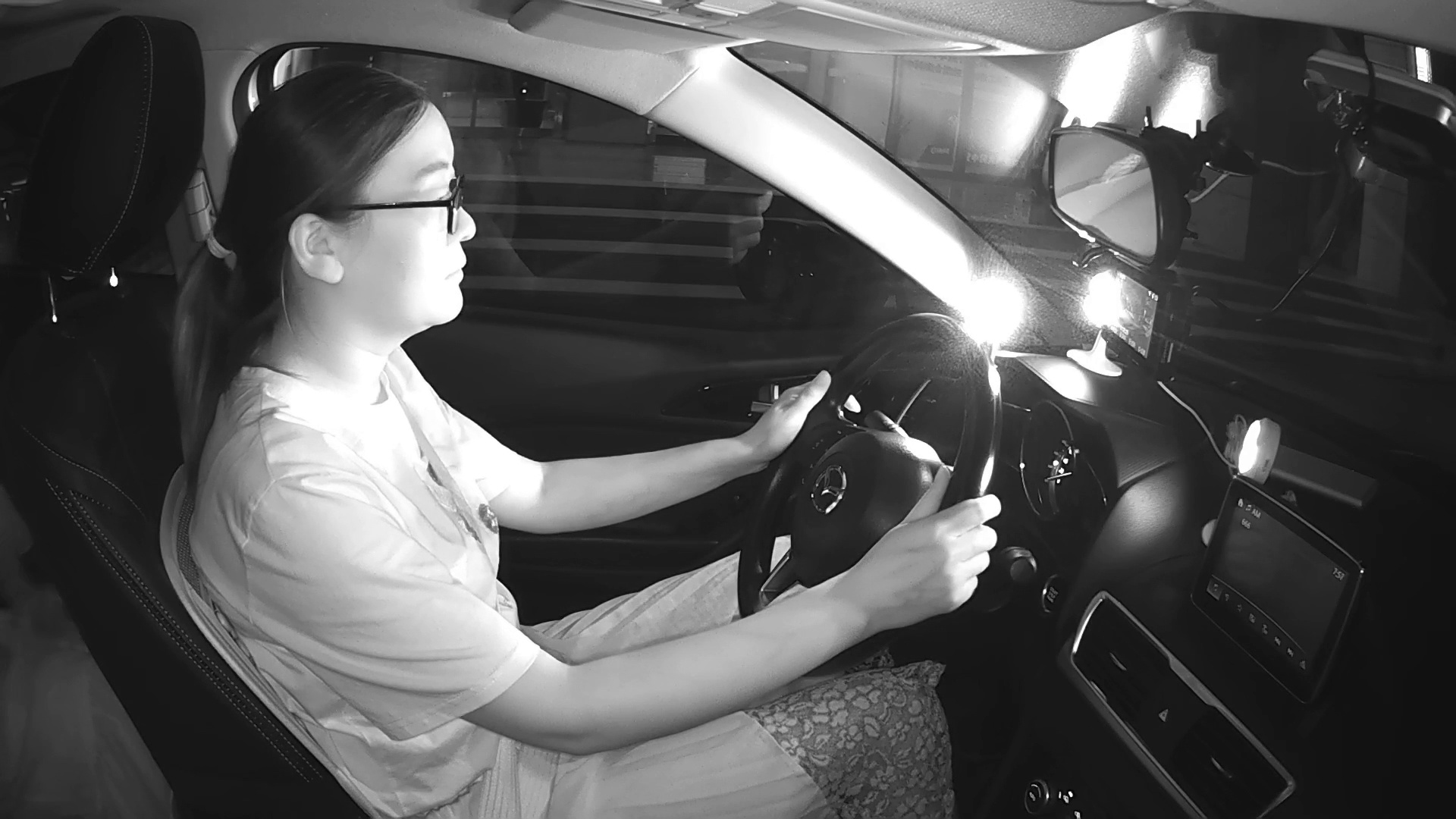} \\
         N1 & N2 & N3 & N4
    \end{tabular}
    \caption{Sample images showcasing a driver engaged in both safe and distracted driving behavior, taken from the 100-Driver dataset \cite{wang2023100}. The images are captured with four cameras installed in the same vehicle in both daytime (D1, D2, D3 and D4) and nighttime (N1, N2, N3 and N4) contexts.}
    \label{fig:samples}
\end{figure*}

The driving sessions were recorded in both daytime and nighttime in five different vehicles, comprising two sedans, two SUVs, and one van. Within each vehicle, four cameras were strategically placed - on the left front, right front, and right side of the drivers. This arrangement served dual purposes. Firstly, having a dataset with multiple cameras aids in developing more robust models that can account for variations in camera perspectives, thus evaluating the models' generalization across different cameras. Secondly, the multi-camera dataset offers an opportunity to enhance system performance by considering the content captured from multiple camera angles. Including nighttime recordings is also essential, as distraction-related behaviors -- such as eye closure and yawning -- are more frequent and harder to detect under low-light conditions. These recordings allow us to assess model robustness across varying lighting environments. Figure \ref{fig:samples} reports samples from the four cameras in both day and night times.

We also perform cross-dataset evaluation on the AUCDD-V1 \cite{eraqi2019driver}, EZZ2021 \cite{ezzouhri2021robust} and SFD \cite{statefarm2020sfd} datasets. Image for these datasets are annotated based on 10 different driver actions: Safe/Normal driving, Texting (right), Talking with cellphone (right), Texting (left), Talking with cellphone (left), Adjusting radio, Drinking, Reaching behind, Hair and makeup, Talking to passenger. AUCDD-V1 consists of 12,978 training images and 4,331 test images. We evaluate model performance only on the test images. In the EZZ2021 dataset, 9 drivers are depicted performing the previous actions. Since a data split is not provided, all images are used for testing. The SFD dataset, used in a Kaggle competition, comprises 22,424 training images and 79,726 test images. The training images are labeled with 10 class labels, while the test images are not labeled. Therefore, only the training set is used for evaluation.

\subsection{Experimental Setup}

\subsubsection{Evaluation}
We run experiments in a cross-camera setup which is implemented by applying the leave-one-camera-out protocol on the 100-Driver dataset. Given that the cameras were placed at 4 different points, for each round 3 cameras are used for training and the other one for evaluation. Our train-val-test splits are obtained by aggregating the splits defined in \cite{wang2023100} according to our protocol. We measure method performance in terms of Top-1 and Top-5 accuracy.

\subsubsection{Backbones}
We select three efficient backbone models -- GhostNet-v1.0~\cite{han2020ghostnet}, MobileNetV3-S~\cite{howard2019searching}, and SqueezeNet-v1.1~\cite{iandola2016squeezenet} -- to evaluate the impact of varying parameter counts on our method. These models are widely recognized as lightweight and computationally efficient, making them well-suited for on-edge deployment in embedded automotive systems~\cite{liu2024image,Baheti_Computationally2020}.

\subsubsection{Implementation details}
Our experiments are performed on a computer featuring an Intel i7-8700 CPU @3.20GHz and an NVIDIA GeForce GTX 1070 Ti. The operating system is Ubuntu 22.04, and the framework is PyTorch 2.0.1. We utilize the SGD optimizer with a momentum of 0.9 and a weight decay equal to $5 \times 10^{-4}$. The batch size is set to 64. The GhostNet-v1.0 backbone is pretrained on ImageNet. The view queries are randomly initialized by sampling from a normal distribution with mean equal to 0 and variance 1. The two hyperparameters in Eq. \ref{eq:loss}, $\lambda_{\rm{ac}}$ and $\lambda_{\rm{vc}}$, are both empirically set to 1.

The learning rate is initialized to 0.01 and reduced by a factor of 10 at 10 and 30 epochs. The inputs are first resized to 224 × 224 pixels. We then use random rotation in the range [-30$^\circ$, 30$^\circ$], random perspective, random color transformations in terms of contrast, saturation and hue, Gaussian blur, and random erasing \cite{zhong2020random} for data augmentation. Training is performed for a total of 50 epochs. The best-performing epoch on the validation set in terms of Top-1 accuracy is then chosen for evaluation on the test data. The margin $\delta$ of triplet losses is empirically set to 1. Performance metrics are Top-1 and Top-5 accuracy.

\subsection{Results}

\subsubsection{Ablation study}
In this section, we present the results highlighting the contribution of the key components of the proposed model. Specifically, we compare three different efficient backbones, assess the contribution of contrastive learning, evaluate the effect of hyperparameter optimization for balancing the two triplet losses, examine the influence of the disentanglement module, and finally, analyze the effect of integrating all component together.

\paragraph{Model backbone} Table \ref{tab:architectures} reports the comparison of the three model architectures in terms of number of parameters and number of floating point operations per second (FLOPS). The GhostNet-v1.0 model has 3.9 million parameters and 0.16 GFLOPS, which indicates a moderate computational complexity with a relatively higher number of parameters compared to the other models. On the other hand, the MobileNetV3-S model, demonstrates a significantly lower number of parameters at 2.5 million and the lowest computational complexity with 0.06 GFLOPS. Finally, the SqueezeNet-v1.1 model features the smallest number of parameters at 1.2 million but has the highest computational complexity among the three, with 0.35 GFLOPS.
\begin{table}
    \centering
    \caption{Comparison of the three models exploited as backbone of the proposed method in terms of number of parameters and computational complexity.}
    \label{tab:architectures}
    \begin{tabular}{lcc}
    \toprule
         Model & Params & GFLOPS \\ \midrule
         GhostNet-v1.0~\cite{han2020ghostnet} & 3.9M & 0.16 \\
         MobileNetV3-S~\cite{howard2019searching} & 2.5M & 0.06 \\
         SqueezeNet-v1.1~\cite{iandola2016squeezenet} & 1.2M & 0.35 \\ \bottomrule
    \end{tabular}
\end{table}

The results reported in Table \ref{tab:backbone-results} reveal notable performance differences among the considered backbones in the leave-one-camera-out experiments. GhostNet consistently outperforms the others across most daytime and nighttime scenarios, achieving, for example, a Top-1 accuracy of 62.51\% and Top-5 accuracy of 93.75\% on D2. This strong performance suggests that GhostNet is particularly effective in handling variations across different camera configurations compared to the other architectures. This aligns with its design goals -- GhostNet was specifically developed to overcome the limitations of earlier architectures, such as MobileNet and SqueezeNet, by increasing feature diversity through the Ghost module while maintaining computational costs low. MobileNetV3 also performs well -- particularly under nighttime conditions, with Top-1 accuracy of 33.18\% for N1 and 47.75\% for N2 -- indicating its robustness in lower light conditions. However, its overall performance remains slightly lower than GhostNet's across most evaluations maybe due to its aggressive optimization for efficiency. SqueezeNet, although highly compact, struggles significantly in challenging scenarios such as D4 and N4 (Top-1 accuracy of 7.91\% and 4.22\%, respectively), likely due to its limited feature diversity resulting from aggressive parameter reduction.
\begin{table*}
    \centering
    \caption{Comparison of the considered backbones in terms of top-1 and Top-5 (\%) accuracy for our leave-one-camera-out experiments. D and N indicate acquisitions during the daytime and nighttime, respectively. The number following the D or N indicates the camera ID. The names on the left side of the arrow refer to the training cameras, while the name on the right side of the arrow refers to the evaluated camera.}
    \label{tab:backbone-results}
    \resizebox{\textwidth}{!}{\begin{tabular}{l|cc|cc|cc|cc|cc|cc|cc|cc}
    \toprule
            & \multicolumn{2}{c}{D\{2,3,4\} $\rightarrow$ D1} & \multicolumn{2}{c}{D\{1,3,4\} $\rightarrow$ D2} & \multicolumn{2}{c}{D\{1,2,4\} $\rightarrow$ D3} & \multicolumn{2}{c|}{D\{1,2,3\} $\rightarrow$ D4} & \multicolumn{2}{c}{N\{2,3,4\} $\rightarrow$ N1} & \multicolumn{2}{c}{N\{1,3,4\} $\rightarrow$ N2} & \multicolumn{2}{c}{N\{1,2,4\} $\rightarrow$ N3} & \multicolumn{2}{c}{N\{1,2,3\} $\rightarrow$ N4} \\
            & Top-1 & Top-5 & Top-1 & Top-5 & Top-1 & Top-5 & Top-1 & Top-5 & Top-1 & Top-5 & Top-1 & Top-5 & Top-1 & Top-5 & Top-1 & Top-5 \\ \midrule
        GhostNet & \textbf{32.37} & \textbf{72.55} & \textbf{62.51} & \textbf{93.75} & \textbf{50.99} & \textbf{82.56} & \textbf{11.29} & \textbf{39.31} & 31.95 & 69.09 & \textbf{49.19} & \textbf{83.79} & \textbf{19.39} & \textbf{55.85} & \textbf{8.63} & \textbf{29.69} \\
        MobileNetV3 & 18.72 & 63.19 & 53.48 & 85.45 & 37.23 & 72.22 & 9.10 & 38.60 & \textbf{33.18} & \textbf{70.21} & 47.75 & 81.78 & 18.37 & 49.68 & 5.30 & 24.25 \\
        SqueezeNet & 15.72 & 47.01 & 35.62 & 82.39 & 39.11 & 72.90 & 7.91 & 32.54 & 33.02 & 69.38 & 45.99 & 78.42 & 15.84 & 46.27 & 4.22 & 21.06 \\ \bottomrule
    \end{tabular}}
\end{table*}

\paragraph{Contrastive learning contribution} This version of DBMNet consists of the Backbone, and the two classifiers, namely the View and Action classifiers. The model is optimized by combining cross-entropy losses and triplet losses for each task. The results for this version are reported in the first row of Table \ref{tab:ablation_results}. As it is possible to see, the lack of the feature disentanglement module compared to the full solution (the results of which are shown in the last row of the table) negatively impacts the results. On average, there is a loss of 5.42\% and 4.05\% on Top-1 and Top-5 accuracy, respectively. The highest gap corresponding to 13.54\% of Top-1 accuracy is registered for Camera 3.

\begin{figure}
    \centering
    \includegraphics[width=\columnwidth]{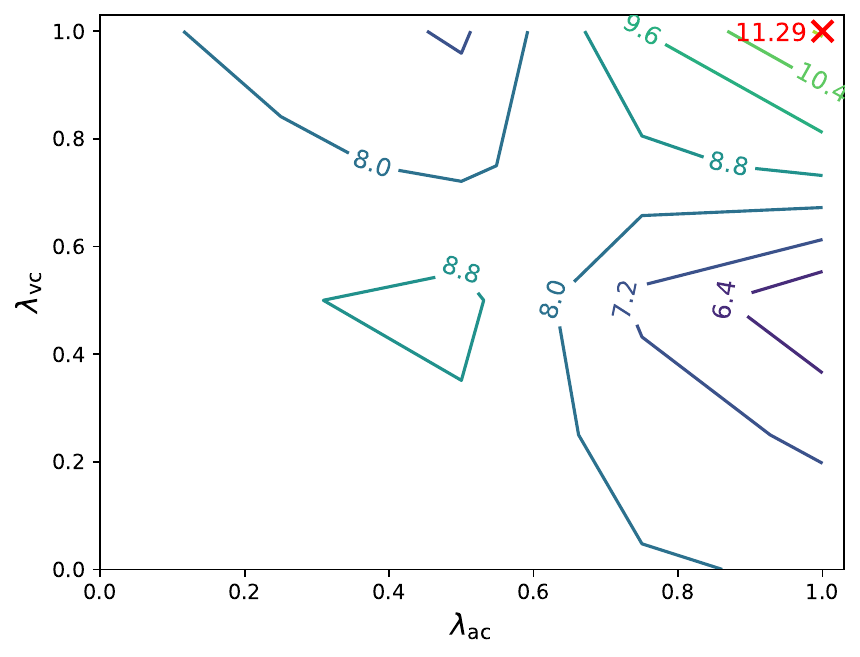}
    \caption{Accuracy for the model trained on D1, D2, and D3 and tested on D4, as a function of the hyperparameters $\lambda_{\rm{ac}}$ and $\lambda_{\rm{vc}}$ defined in the Eq. \ref{eq:loss}. The red cross indicates the optimal configuration that yields the highest accuracy.}
    \label{fig:triple-loss-hyper-tuning}
\end{figure}

\paragraph{Triplet loss hyperparameter tuning}
The balance between the two triplet losses -- $\mathcal{L}_{\rm{ac}}$ and $\mathcal{L}_{\rm{vc}}$ -- plays a critical role for optimizing model performance. In this experiment, we assess Top-1 accuracy as a function of the weighting parameters $\lambda_{\rm{ac}}$ and $\lambda_{\rm{vc}}$ as defined in Eq. \ref{eq:loss}, by training on D1, D2, and D3 and testing on D4. As shown in Figure \ref{fig:triple-loss-hyper-tuning}, the highest accuracy of 11.29\% is achieved when both losses are equally weighted with $\lambda_{\rm{ac}} = \lambda_{\rm{vc}} = 1$. We can also observe that performance slightly degrades when the weights of the two losses are similar but lower than one. In contrast, performance significantly degrades when the balance between the two losses is skewed, with the lowest accuracy of 5.68\% observed at $\lambda_{\rm{ac}} = 1$ and $\lambda_{\rm{vc}} = 0.5$.

\paragraph{Feature disentanglement module contribution} In this experiment, the architecture of DBMNet remains unchanged. The difference with the complete version lies solely in the choice of loss function utilized for model optimization. Specifically, we incorporate only the two cross-entropy losses. The outcomes for this variant are depicted in the second row of Table \ref{tab:ablation_results}. We emphasize that also the performance of this solution is inferior to that of the full version of DBMNet. Notably, there is a 6.94\% drop in Top-1 and 8.79\% drop in Top-5. More significantly, these results are even lower than those achieved with Contrastive learning alone. Consequently, it is evident that while the Feature Disentanglement module offers benefits to DBMNet, the use of Contrastive learning yields even greater advantages.
\begin{table*}
    \centering
    \caption{Ablation experiments about feature disentanglement and contrastive learning, using Top-1 and Top-5 (\%) accuracy for leave-one-camera-out experiments. The results in the first row of this table match those reported for GhostNet \cite{wang2023100} in Table \ref{tab:comparison-results} This is because GhostNet \cite{wang2023100} was trained using only cross-entropy loss, without the inclusion of feature disentanglement or contrastive learning.}
    \label{tab:ablation_results}
    \begin{tabular}{cc|cc|cc|cc|cc}
        \toprule
            \multirow{2}{*}{\makecell{Feat.\\ disent.}} & \multirow{2}{*}{\makecell{Contr.\\ learn.}} & \multicolumn{2}{c|}{\{D2,D3,D4\} $\rightarrow$ D1} & \multicolumn{2}{c|}{\{D1,D3,D4\} $\rightarrow$ D2} & \multicolumn{2}{c|}{\{D1,D2,D4\} $\rightarrow$ D3} & \multicolumn{2}{c}{\{D1,D2,D3\} $\rightarrow$ D4} \\
            & & Top-1 & Top-5 & Top-1 & Top-5 & Top-1 & Top-5 & Top-1 & Top-5 \\ \midrule
            & & 21.54 & 64.96 & 57.89 & 85.46 & 33.79 & 66.12 & 8.85 & 32.35 \\
            & \checkmark & 28.99 & 70.00 & 61.05 & 91.58 & 37.45 & 76.32 & 7.99 & 34.06 \\
            \checkmark & & 28.22 & 66.15 & 58.13 & 87.47 & 34.27 & 65.87 & 8.80 & 33.51 \\
            \checkmark & \checkmark & \textbf{32.37} & \textbf{72.55} & \textbf{62.51} & \textbf{93.75} & \textbf{50.99} & \textbf{82.56} & \textbf{11.29} & \textbf{39.31} \\
        \bottomrule
    \end{tabular}
\end{table*}
\paragraph{Full solution analysis} 
DBMNet capitalizes on the advantages offered by both the Feature disentanglement module and Contrastive learning, as outlined in the previous sections. Our focus is on another important aspect of our method in this section. Given that DBMNet performs feature disentanglement by highlighting the features for the driver's actions based on the view that is most similar to the test view, it is imperative to ensure the effectiveness of view classification. We register an average performance for the four rounds of leave-one-camera-out equal to 98.54\% on the validation set of Driver-100. This means that the view classification is accurate and should be able to correctly guide the choice of the view queries.

\subsubsection{Comparison with other methods}
In this section, we compare the results achieved by our DBMNet with those of seven recent methods, namely DGCCL \cite{yang2025domain}, Duan \etal~\cite{duan2023enhancing}, GhostNet \cite{wang2023100}, Hu \etal~\cite{Hu_Driver2023}, MambaVision~\cite{hatamizadeh2025mambavision}, MobileViT~\cite{mehtamobilevit}, OLCMNet \cite{Li_Driver2022}, Si-CA MobileNet \cite{lv2025si}, and SwinT~\cite{liu2021swin}. DGCCL is based on a ViT-B/16 architecture and is optimized using a composite loss function that combines center loss and ArcFace loss to enhance domain generalization. It is worth noting that Duan \etal~trained a GhostNet-v1.0 by exploiting their proposed S-Softmax classifier for improving model robustness. GhostNet is the same architecture used in our method as the backbone, and unlike our method, it is trained exclusively through cross-entropy loss optimization for driver action classification. Hu \etal~\cite{Hu_Driver2023} trained a ResNet-18 using a compound loss, which includes the proposed Clustering Supervised Contrastive Loss (C-SCL) for improving action representations, along with the cross-entropy loss used for the classification head. MambaVision is a hybrid Mamba-Transformer backbone optimized for efficient visual feature modeling with self-attention for global context, and we adopt its Tiny version for our experiments. MobileViT combines the efficiency of lightweight convolutional neural networks with the representational power of Transformers by embedding ViT blocks into a mobile-friendly CNN backbone, enabling strong global context modeling with low computational cost. OLCMNet is a lightweight network with an OLCM block to improve model effectiveness. Finally, Si-CA MobileNet is a lightweight CNN for distracted driver detection that integrates a parameter-free attention mechanism into Si-Blocks and Coordinate Attention into CA-Blocks to enhance feature extraction while maintaining efficiency. SwinT, the tiny variant of the Swin Transformer, employs shifted window-based self-attention to capture hierarchical visual representations with reduced complexity, achieving an effective balance between performance and computational efficiency.
\begin{table}
    \centering
    \caption{Comparison of the proposed DBMNet with literature methods in terms of number of parameters and computational complexity.}
    \label{tab:comparison-complex}
    {\begin{tabular}{lcc}
    \toprule
     & Params & GFLOPS \\ \midrule
         DGCCL \cite{yang2025domain} & 85.8M & 17.58 \\
         Duan \etal~\cite{duan2023enhancing} & 4.3M & 0.16 \\
         GhostNet \cite{wang2023100} & 3.9M & 0.16 \\
         Hu \etal~\cite{Hu_Driver2023} & 11.2M & 1.83 \\
         MambaVision \cite{hatamizadeh2025mambavision} & 31.8M & 4.40 \\
         MobileViT~\cite{mehtamobilevit} & 5.6M & 0.70 \\
         OLCMNet \cite{Li_Driver2022} & 2.8M & 0.68 \\
         Si-CA MobileNet \cite{lv2025si} & 3.2M & 0.17\\
         SwinT~\cite{liu2021swin} & 28.3M & 4.49 \\
         DBMNet (ours) & 4.0M & 0.16 \\ \bottomrule
    \end{tabular}}
\end{table}

Tables \ref{tab:comparison-complex} and \ref{tab:comparison-results} present a comparison between the proposed DBMNet and state-of-the-art methods in terms of model complexity and performance. Table \ref{tab:comparison-complex} highlights that DBMNet achieves a favorable balance between accuracy and efficiency, with only 4.0M parameters and 0.16 GFLOPS -- comparable to lightweight models like GhostNet (3.9M, 0.16 GFLOPS) and Si-CA MobileNet (3.2M, 0.17 GFLOPS), while being significantly more efficient than Hu \etal~(11.2M, 1.83 GFLOPS), MambaVision (31.8M, 4.40 GFLOPS), MobileViT (5.6M, 0.70 GFLOPs), SwinT (28.3M, 4.49 GFLOPs), and especially DGCCL (85.8M, 17.58 GFLOPS).

Performance-wise (Table \ref{tab:comparison-results}), DBMNet consistently delivers competitive results across the leave-one-camera-out protocol, ranking among the top two methods in 6 out of 8 test scenarios. The lowest Top-1 accuracy for DBMNet is observed on camera D4 (11.29\%), and the highest on camera D2 (62.51\%). This variation is expected due to the distinct nature of D4 compared to the other cameras (see Figure \ref{fig:samples}). Ranking the cameras by Top-1 accuracy yields the order: D2, D3, D1, and D4 -- underscoring the influence of viewpoint on driver behavior recognition. Compared to DGCCL, which remains the most computationally expensive method by a wide margin, DBMNet achieves comparable or superior accuracy in most cases while being over 20 times more efficient in FLOPS and more than 21 times smaller in parameter count. Specifically, DBMNet outperforms DGCCL in 5 out of 8 test scenarios, with notable gains on N1 (+14.86\% Top-1), N3 (+2.30\% Top-1), and N4 (+3.74\% Top-1). MambaVision and SwinT achieve strong Top-1 accuracy and in some cases surpass DBMNet (e.g., D1 and N1 for MambaVision; D1, D2, and N1–N4 for SwinT), but this comes at a steep computational cost—SwinT requires over 28M parameters and 4.49 GFLOPs. MobileViT also shows solid performance, with accuracy close to DBMNet in several settings (e.g., N3 and N4), but still requires more than 3.5× the FLOPs and over 40\% more parameters. Si-CA MobileNet, while lightweight, lags behind in accuracy across all settings, highlighting the challenge of maintaining performance under strict efficiency constraints.

Overall, DBMNet delivers strong accuracy with minimal computational overhead, offering a practical and effective solution for cross-view driver state recognition, and proving more balanced than both highly efficient yet less accurate models (like Si-CA MobileNet) and accuracy-oriented but computationally heavy ones (like MambaVision, SwinT, and DGCCL).

\begin{table*}
    \centering
    \caption{Top-1 and Top-5 (\%) accuracy for our leave-one-camera-out experiments. D indicates acquisitions during the daytime. The number following the D indicates the camera ID. The names on the left side of the arrow refer to the training cameras, while the name on the right side of the arrow refers to the evaluated camera. In each column, the best and second best results are marked in boldface and underlined, respectively.}
    \label{tab:comparison-results}
    \resizebox{\textwidth}{!}{\begin{tabular}{l|cc|cc|cc|cc|cc|cc|cc|cc}
    \toprule
            & \multicolumn{2}{c}{D\{2,3,4\} $\rightarrow$ D1} & \multicolumn{2}{c}{D\{1,3,4\} $\rightarrow$ D2} & \multicolumn{2}{c}{D\{1,2,4\} $\rightarrow$ D3} & \multicolumn{2}{c|}{D\{1,2,3\} $\rightarrow$ D4} & \multicolumn{2}{c}{N\{2,3,4\} $\rightarrow$ N1} & \multicolumn{2}{c}{N\{1,3,4\} $\rightarrow$ N2} & \multicolumn{2}{c}{N\{1,2,4\} $\rightarrow$ N3} & \multicolumn{2}{c}{N\{1,2,3\} $\rightarrow$ N4} \\
            & Top-1 & Top-5 & Top-1 & Top-5 & Top-1 & Top-5 & Top-1 & Top-5 & Top-1 & Top-5 & Top-1 & Top-5 & Top-1 & Top-5 & Top-1 & Top-5 \\ \midrule
        DGCCL \cite{yang2025domain} & 33.36 & 65.33 & \underline{68.95} & 84.06 & \textbf{51.48} & 67.37 & 10.78 & 34.63 & 17.09 & 37.08 & 48.87 & 68.35 & 17.09 & 37.08 & 4.89 & 28.47 \\
        Duan \etal~\cite{duan2023enhancing} & 8.86 & 40.92 & 14.40 & 47.39 & 12.61 & 42.46 & 7.25 & 30.17 & 13.78 & 43.53 & 11.62 & 43.09 & 5.85 & 25.83 & 4.82 & 27.80 \\
        GhostNet \cite{wang2023100} & 21.54 & 64.96 & 57.89 & 85.46 & 33.79 & 66.12 & 8.85 & 32.35 & 26.43 & 61.16 & 47.12 & {80.90} & 17.37 & 48.24 & 4.89 & 23.97 \\
        Hu \etal~\cite{Hu_Driver2023} & 27.39 & 70.20 & 59.70 & 89.48 & 43.67 & 76.69 & 5.45 & 29.17 & 29.67 & 62.03 & 47.50 & 80.27 & 17.26 & 45.83 & 4.02 & 25.81 \\
        MambaVision \cite{hatamizadeh2025mambavision} & \underline{37.80} & \underline{78.05} & 62.13 & 92.64 & 49.98 & 81.19 & 10.81 & 34.75 & \underline{43.51} & \underline{75.07} & {49.17} & 77.71 & \underline{19.46} & \underline{61.30} & 7.60 & 28.93 \\
        MobileViT~\cite{mehtamobilevit} & 35.69 & 71.20 & 57.33 & 92.70 & 46.82 & 79.01 & 9.58 & 32.48 & 35.47 & 72.65 & 44.93 & 74.08 & 18.60 & 52.97 & 7.36 & 28.87 \\
        OLCMNet \cite{Li_Driver2022} & 20.51 & 68.97 & 55.71 & 81.35 & 34.03 & 68.91 & 8.48 & 30.17 & 25.50 & 63.51 & 40.64 & 76.81 & 12.77 & 30.91 & 5.54 & 23.93 \\
        Si-CA MobileNet \cite{lv2025si} & 12.62 & 42.43 & 20.23 & 60.53 & 12.79 & 38.51 & 5.35 & 26.78 & 13.63 & 42.08 & 17.99 & 45.26 & 7.24 & 27.69 & 5.04 & 27.53 \\
        SwinT~\cite{liu2021swin} & \textbf{42.02} & \textbf{84.52} & \textbf{72.63} & \textbf{95.68} & 49.69 & \underline{81.85} & \textbf{24.07} & \textbf{48.02} & \textbf{50.73} & \textbf{85.53} & \textbf{52.90} & \textbf{88.20} & \textbf{24.36} & \textbf{68.26} & \textbf{10.32} & \textbf{37.68} \\
        DBMNet (ours) & 32.37 & 72.55 & 62.51 & \underline{93.75} & \underline{50.99} & \textbf{82.56} & \underline{11.29} & \underline{39.31} & 31.95 & {69.09} & \underline{49.19} & \underline{83.79} & {19.39} & 55.85 & \underline{8.63} & \underline{29.69} \\ \bottomrule
    \end{tabular}}
\end{table*}

\begin{figure*}
    \centering
    \begin{tabular}{cccc}
         \includegraphics[width=.22\linewidth]{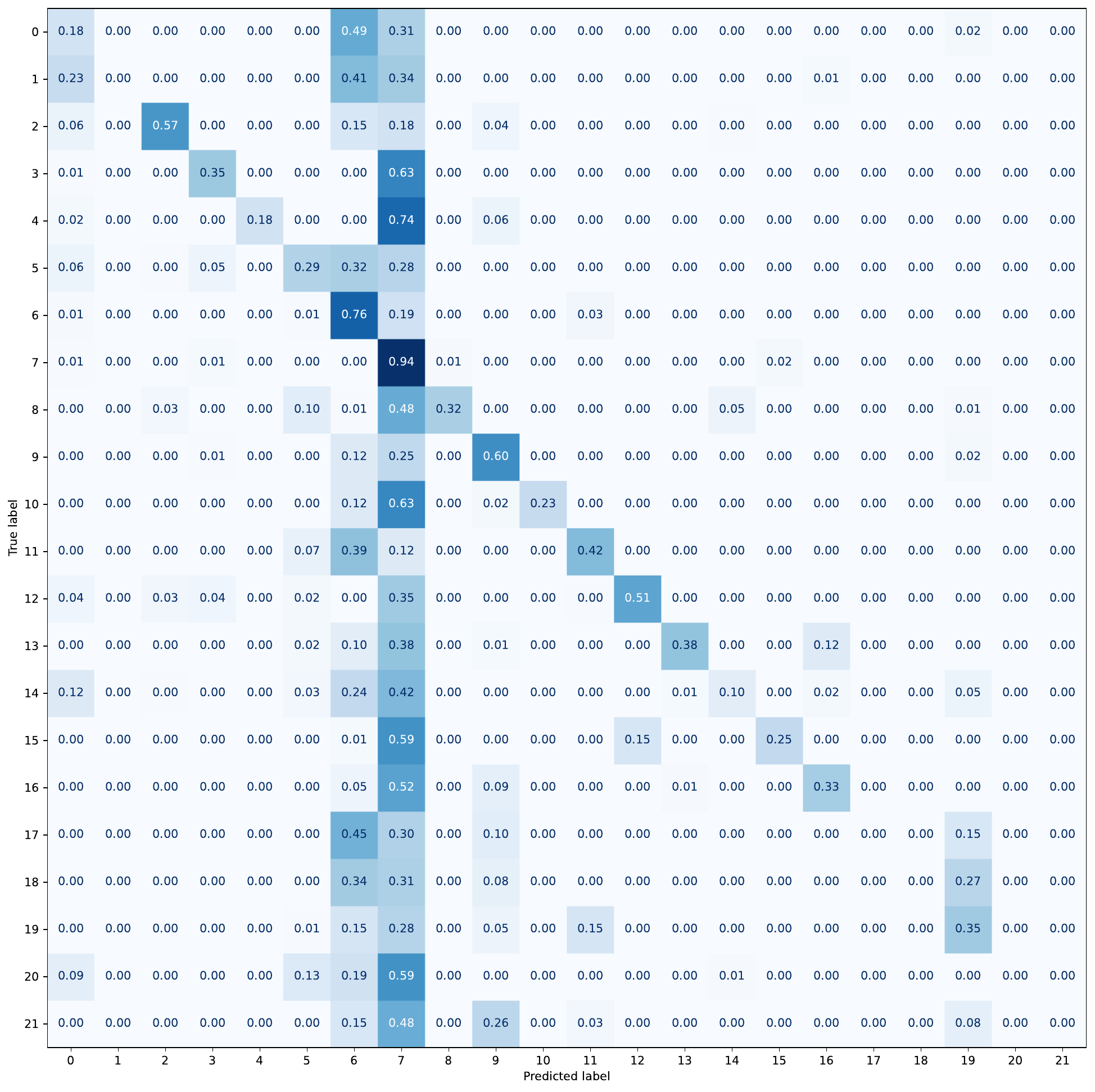} & \includegraphics[width=.22\linewidth]{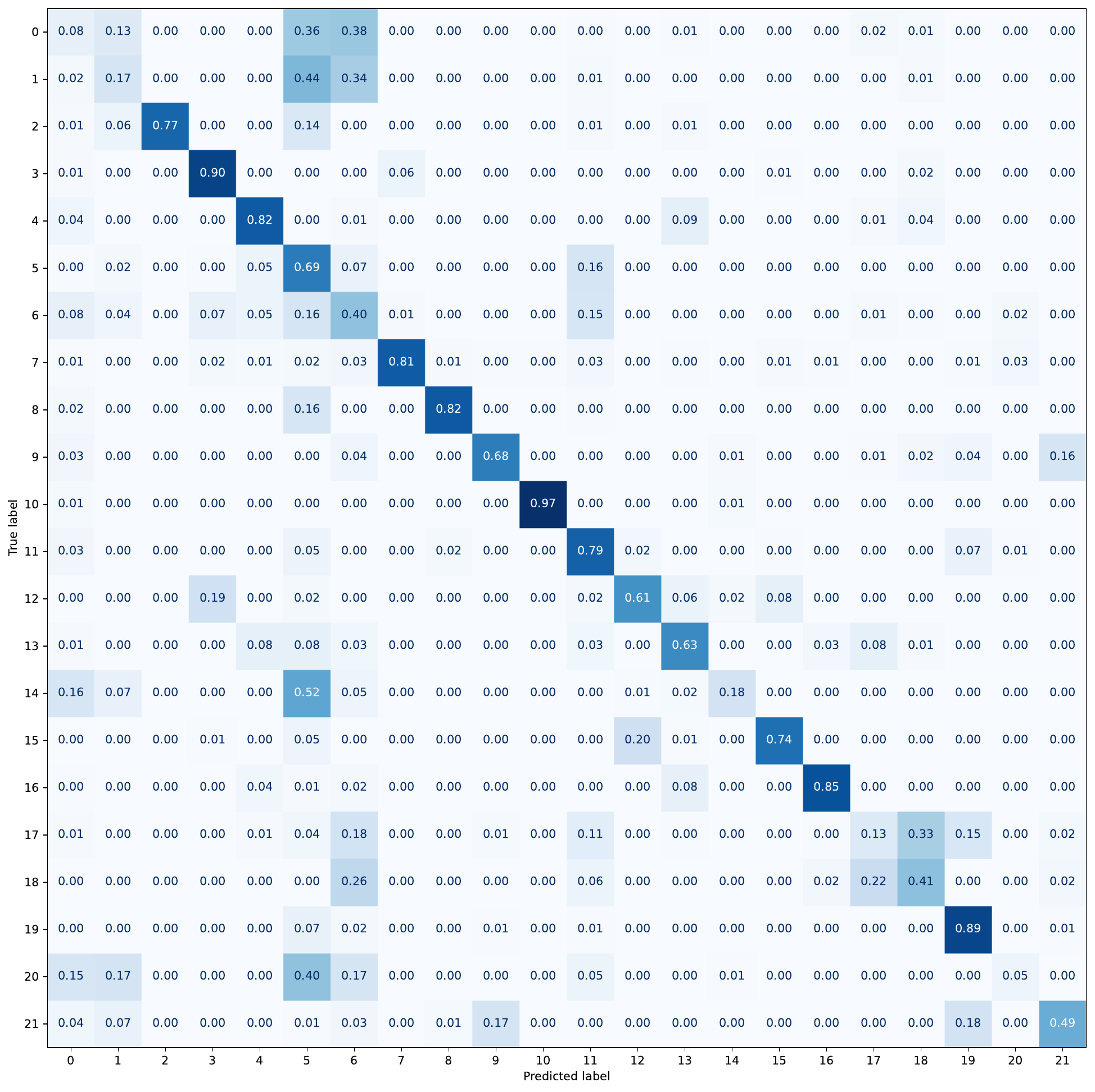} & \includegraphics[width=.22\linewidth]{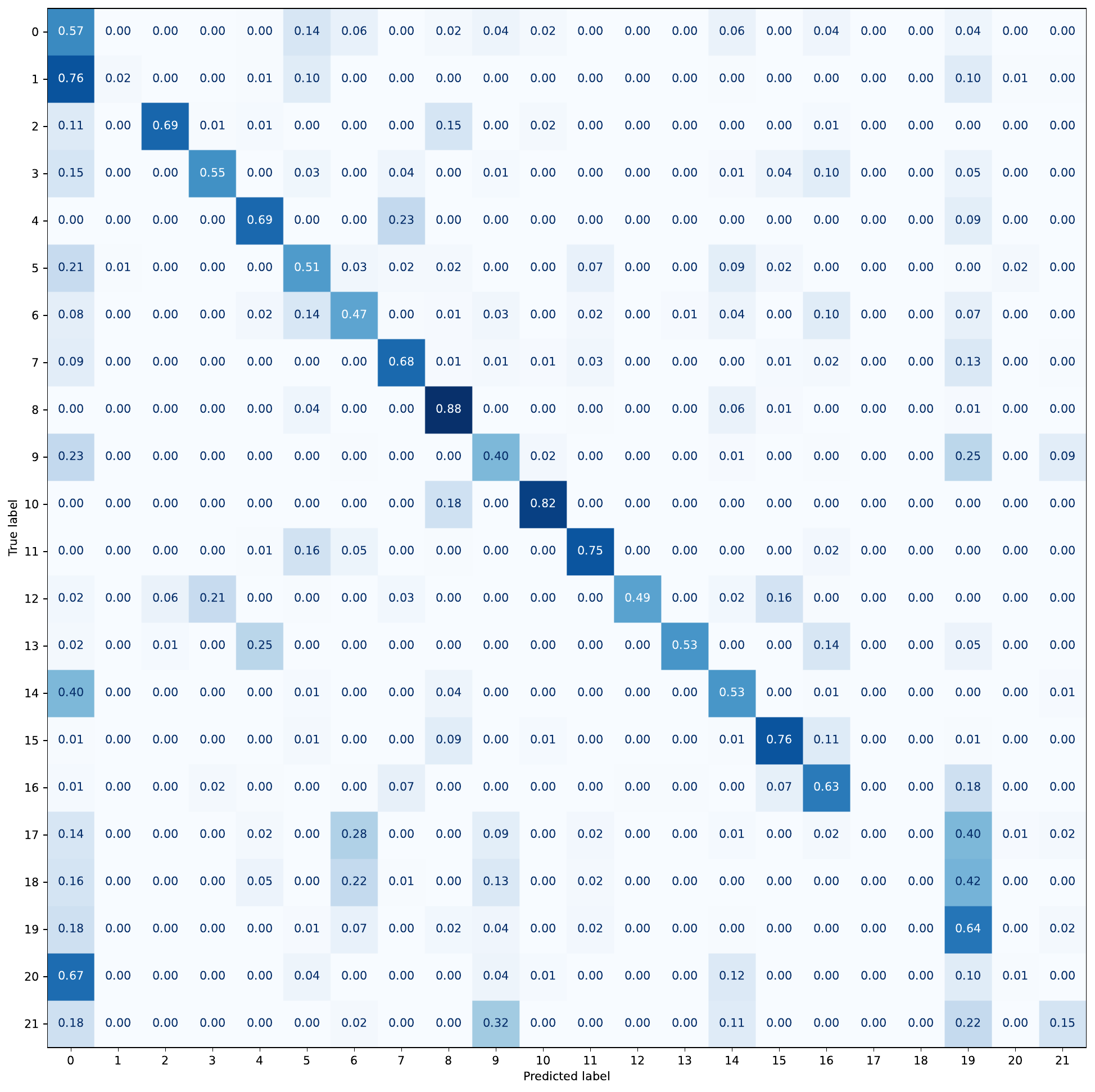} & \includegraphics[width=.22\linewidth]{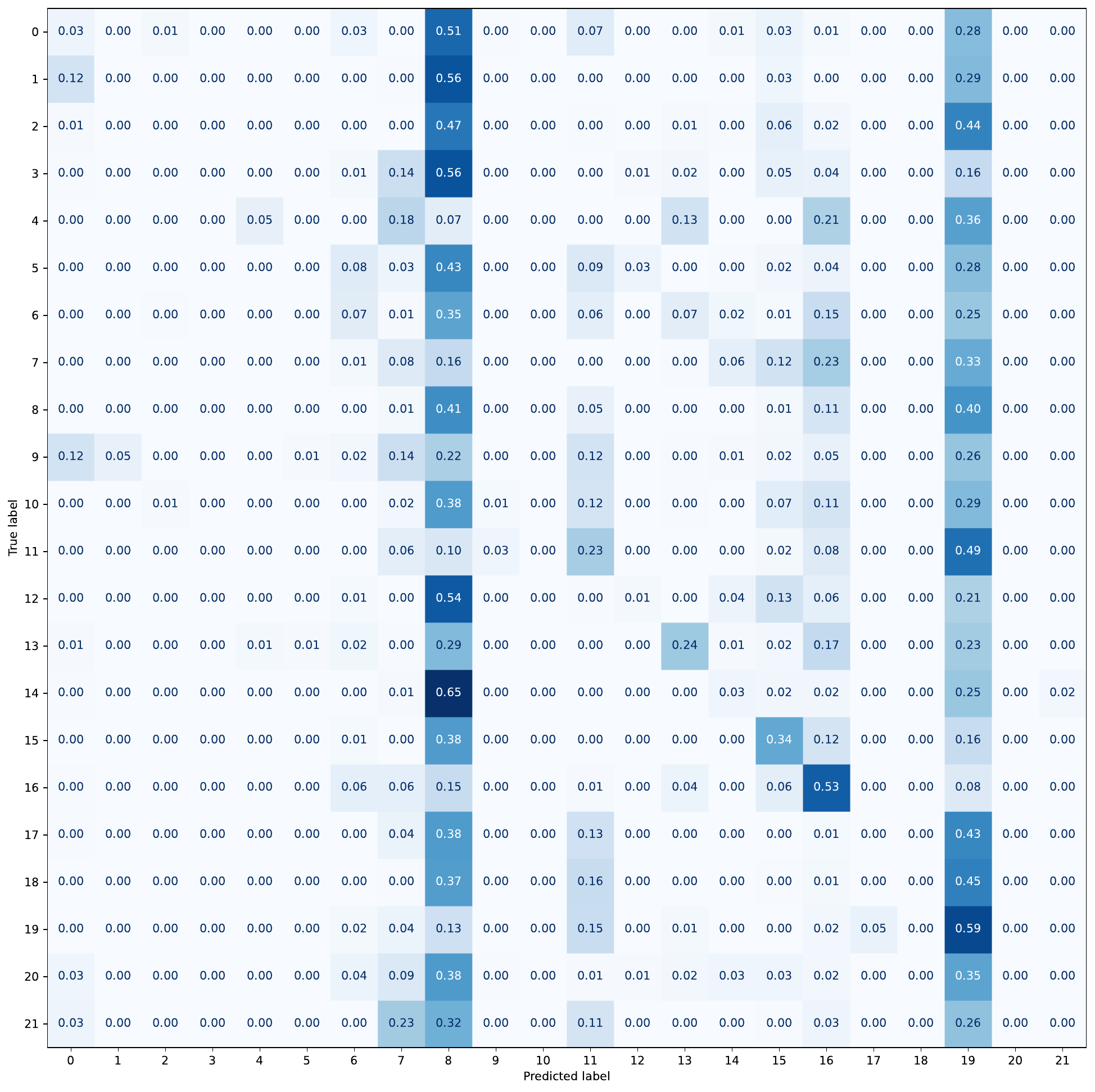} \\
         D1 & D2 & D3 & D4 \vspace{1em} \\
         \includegraphics[width=.22\linewidth]{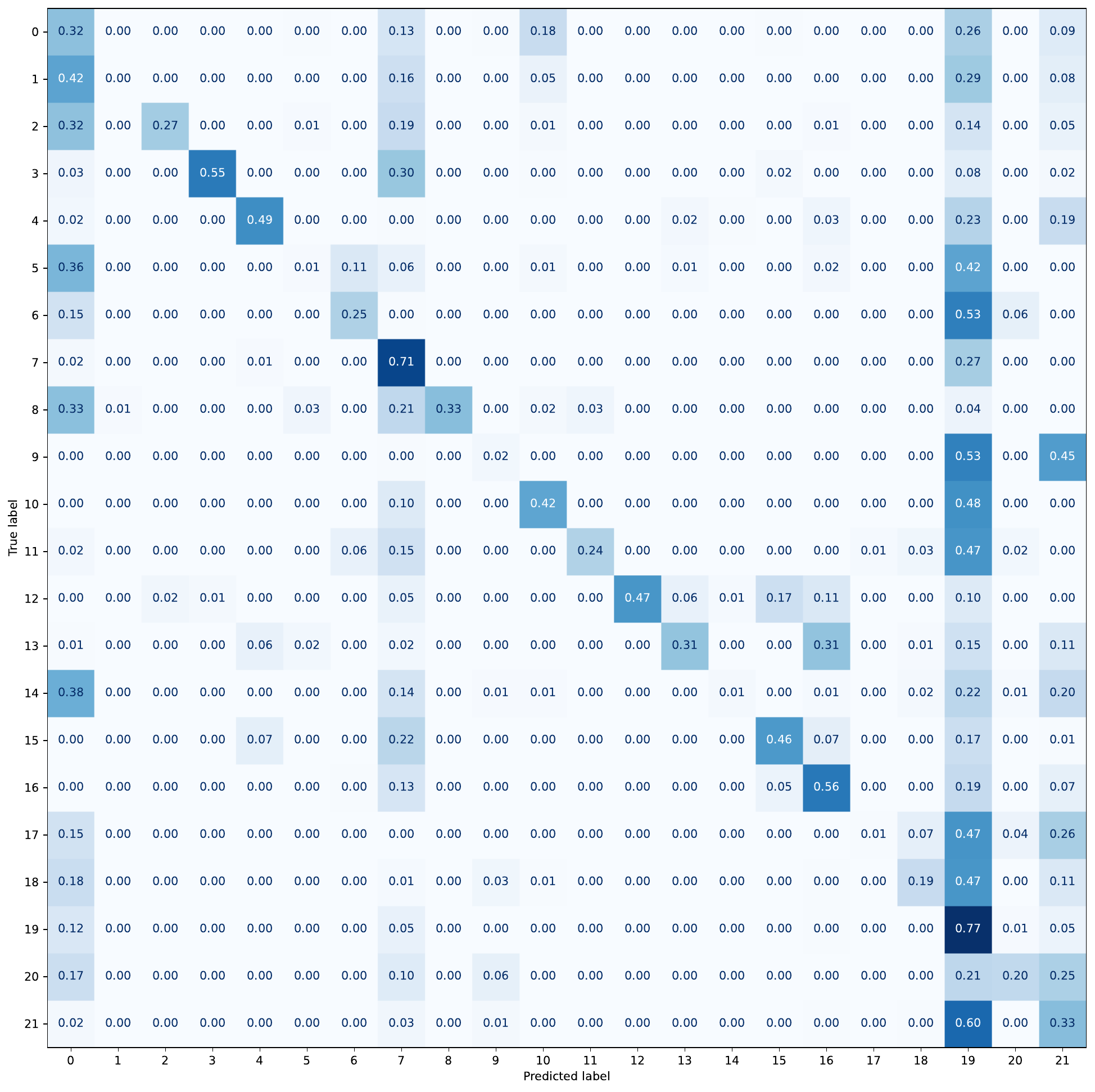} & \includegraphics[width=.22\linewidth]{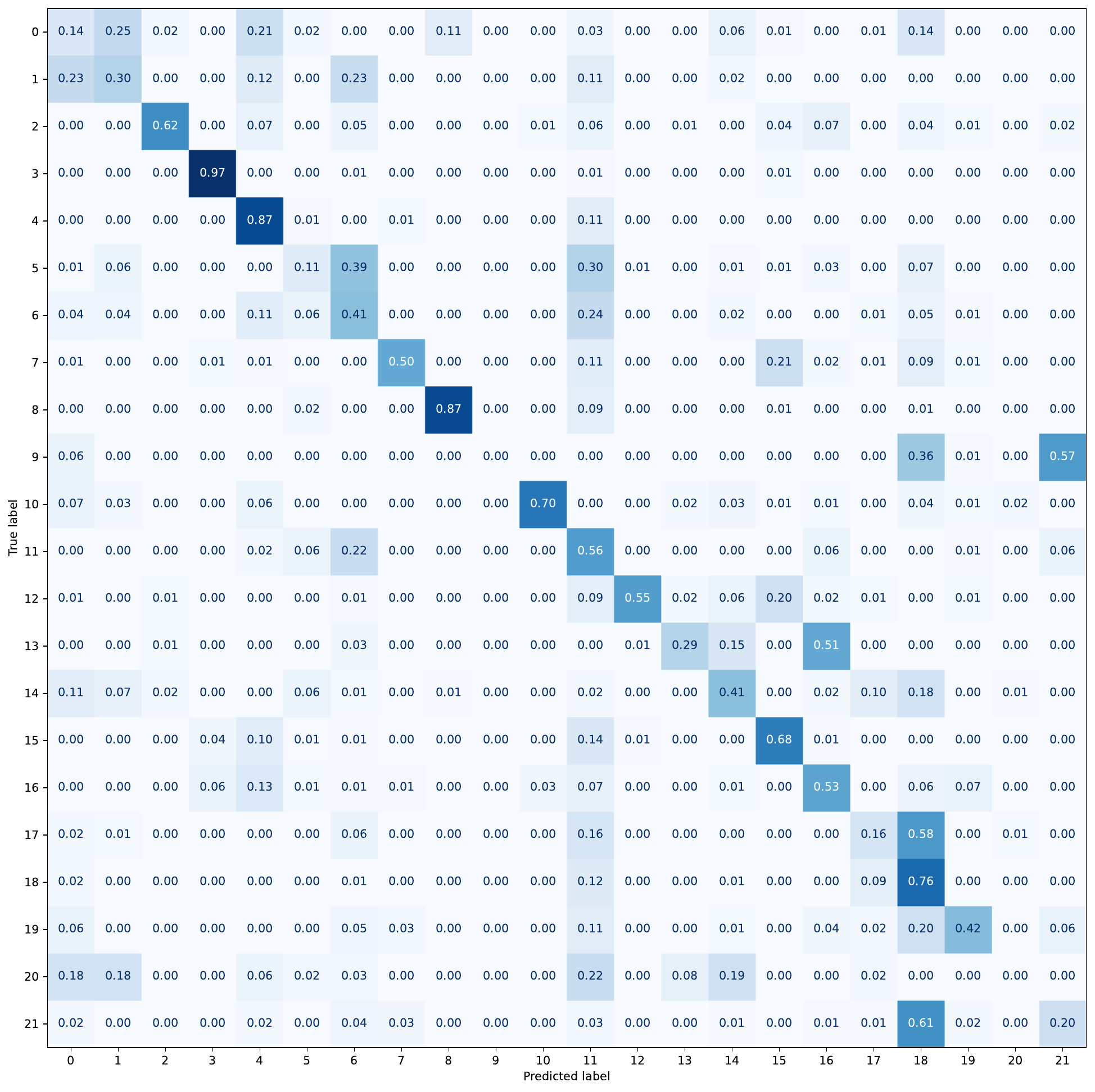} & \includegraphics[width=.22\linewidth]{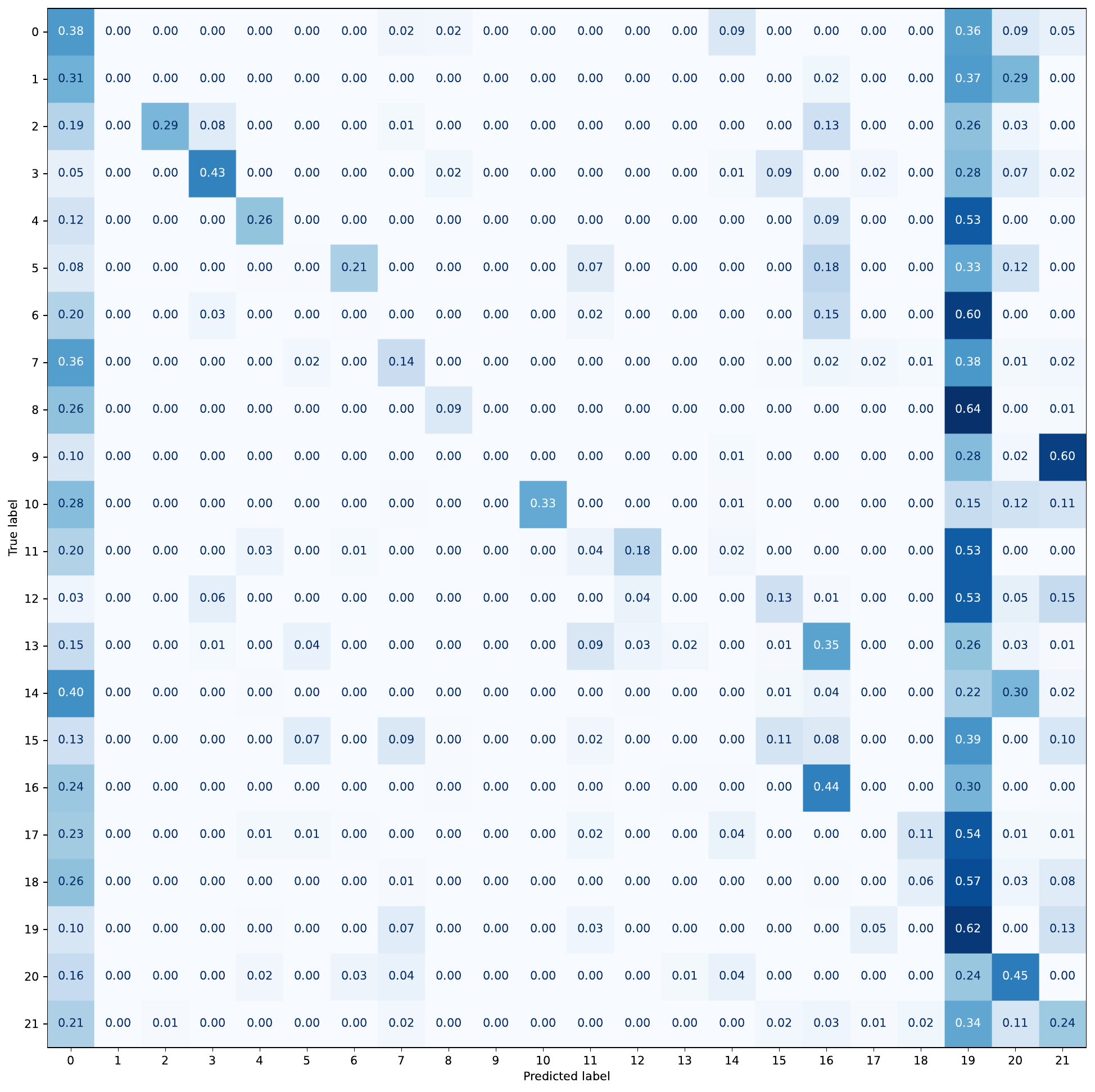} & \includegraphics[width=.22\linewidth]{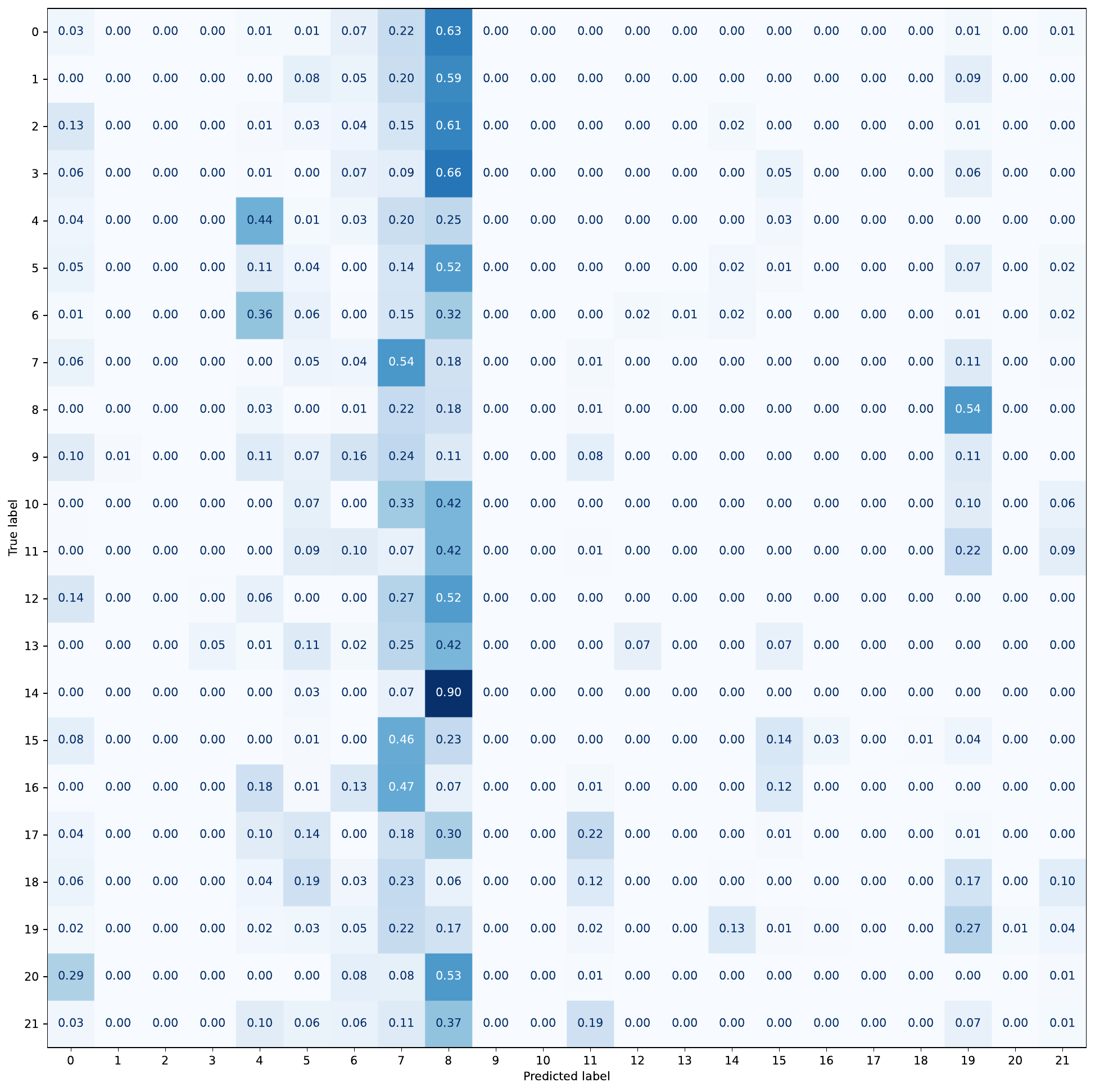} \\
         N1 & N2 & N3 & N4
    \end{tabular}
    \caption{Confusion matrix for each cross-camera configuration (best viewed zoom in).}
    \label{fig:conf-matrix}
\end{figure*}

Figure \ref{fig:conf-matrix} shows the confusion matrices for each cross-camera configuration. For the daytime subset we can the following observations. Confirming that the best performance is obtained for D2 and D3, there is a prominent diagonal in the respective confusion matrices. The confusion matrix for D1 shows a high bias in the class ``Hair / Makeup'' (\#7). Many of the test samples are wrongly categorized with this class. An analysis of the distribution of the samples per category, however, excludes the presence of an imbalance in sample cardinality that could favor any specific category. Therefore, the negative results can only be attributed to the poor generalization ability of the model on this viewpoint. In the case of D4, ``Looking to the left'' (\#8) and ``Reaching behind'' (\#19) consistently emerge as the most frequently predicted categories for the test samples, even when predictions are incorrect. With an accuracy rate of 53.00\%, the class ``Drinking / Eating (right)'' stands as the second most accurately classified category for this camera. In general, for all daytime cameras, the ``Looking Left'' (\#8) class is recognized with the highest rate. Confusion matrices for the nighttime subset reflect the poor performance, with only N1 and N2 presenting a discernible diagonal. The highest performance for N1 is observed in the ``Reaching behind'' (\#19) class with 77\% accuracy and the ``Hair / Makeup'' (\#7) class with 71\% accuracy. For N2, several classes achieve accuracy higher than or equal to 70\%, namely ``Talk with cellphone (left)'' (\#3), ``Talk with cellphone (right)'' (\#4), ``Looking Left'' (\#8), ``Looking up'' (\#10), and ``Operating GPS'' (\#18). In the case of N3, the model shows a high concentration of false positives for the ``Normal driving'' (\#0) and ``Reaching behind'' (\#19) classes. For N4, ``Talk with cellphone (right)'' (\#4) and ``Hair / Makeup'' (\#7) classes are recognized with accuracies of 44\% and 54\%, respectively. The lower performance for N4 might be due to its significant difference from the other views, making it challenging to recognize actions accurately, with only these two actions being relatively easier to identify.

\subsubsection{Cross-dataset results}
\begin{figure*}
    \centering
    \begin{tabular}{cccc}
        \includegraphics[height=60px]{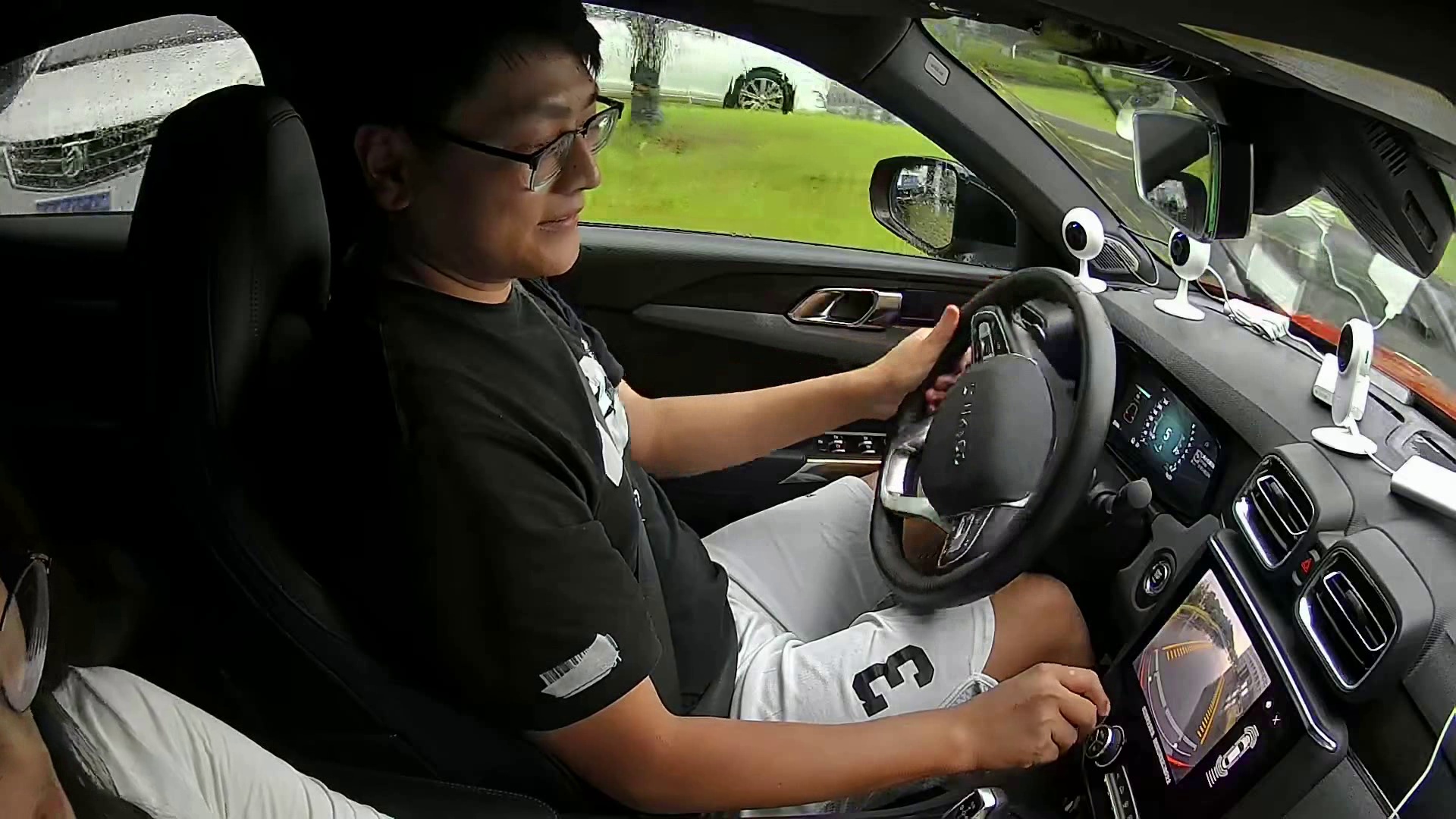} & \includegraphics[height=60px]{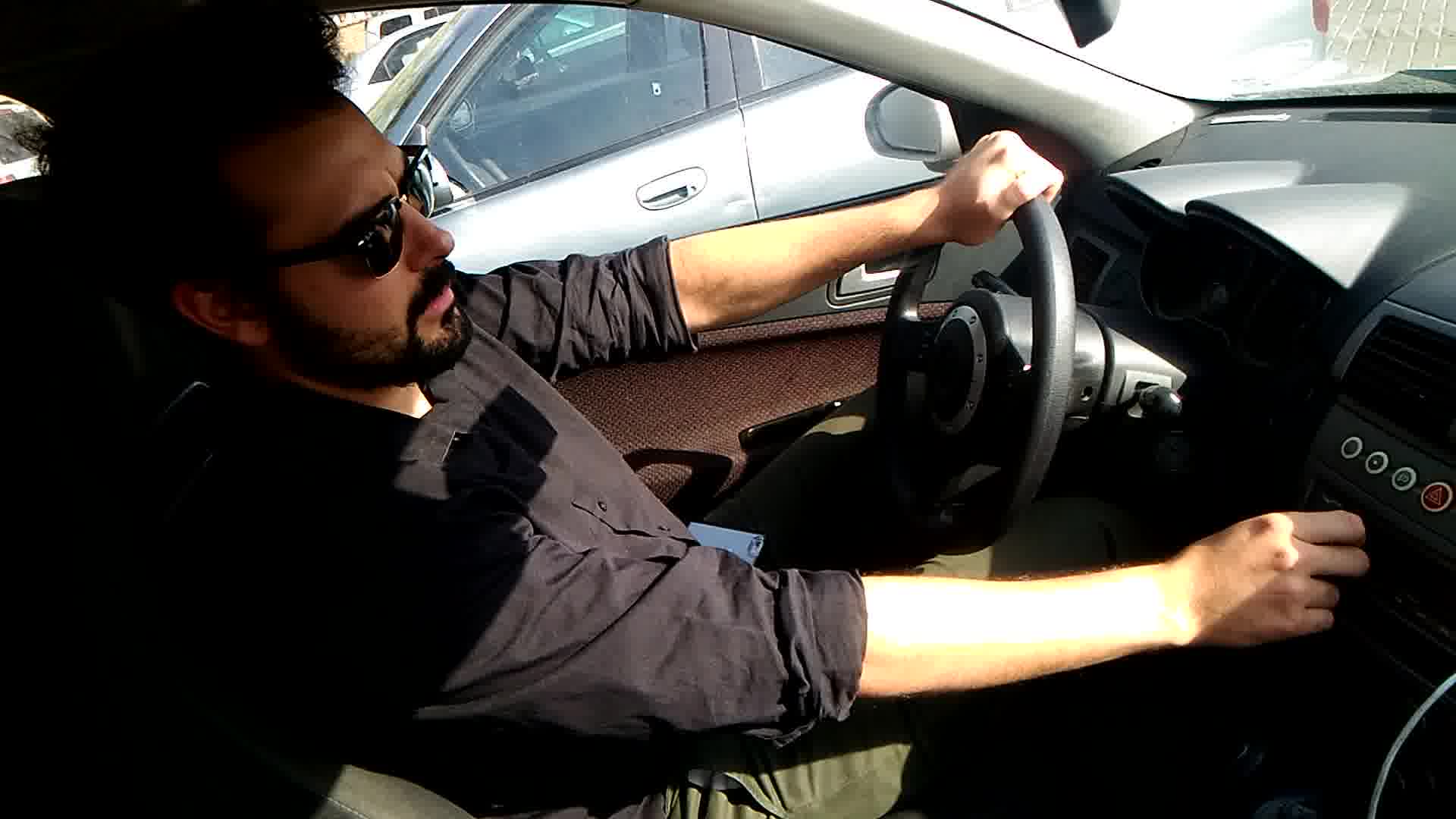} &
        \includegraphics[height=60px]{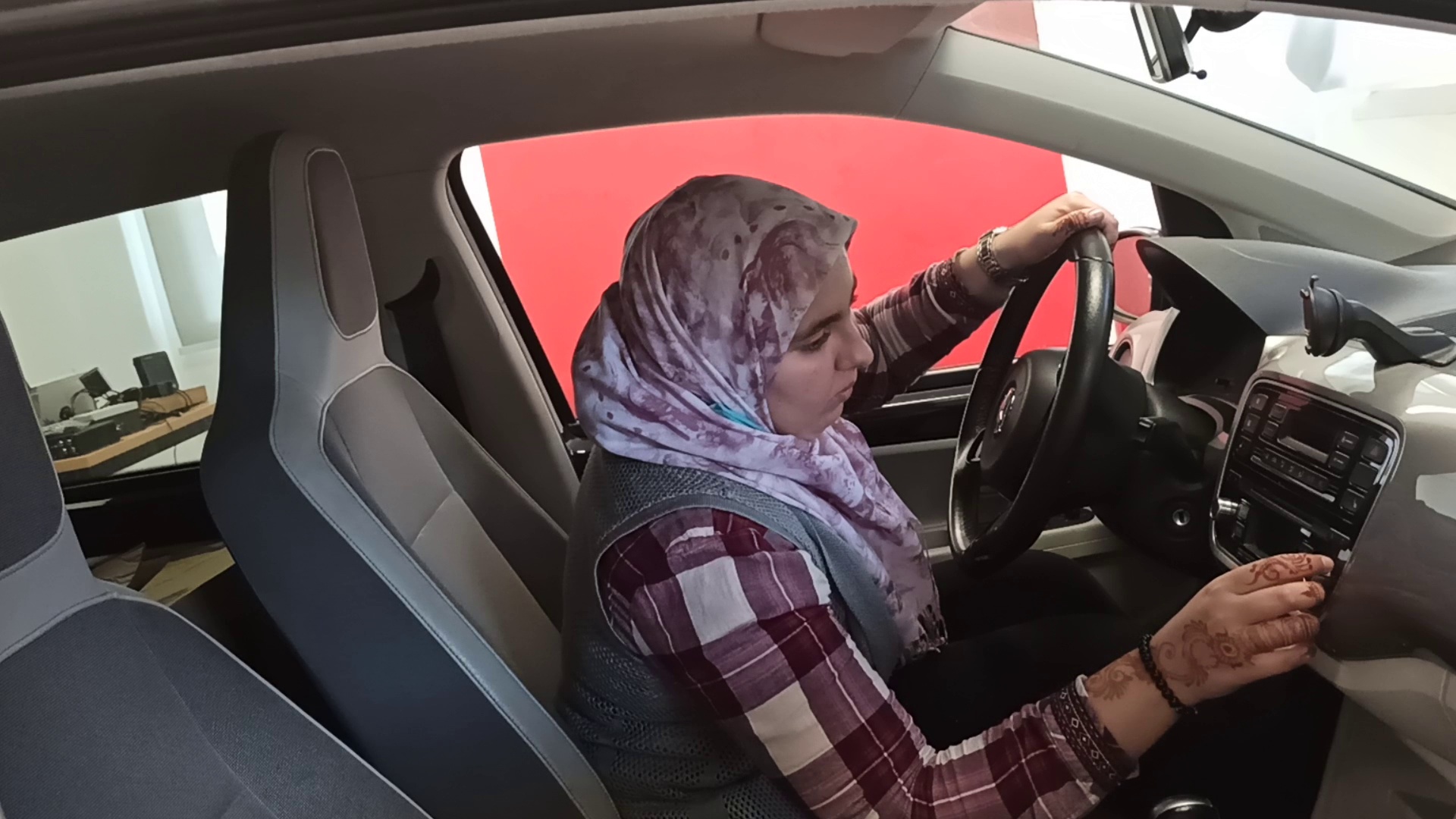} &
        \includegraphics[height=60px]{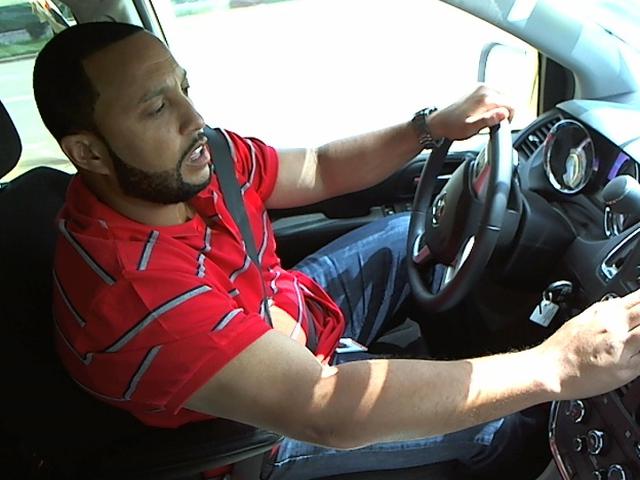} \\
        100-Driver & AUCDD-V1 & EZZ2021 & SFD
    \end{tabular}
    \caption{Sample images showing consistency in framing in the 100-Driver (using the D4 camera) \cite{wang2023100}, AUCDD-V1 \cite{eraqi2019driver}, EZZ2021 \cite{ezzouhri2021robust} and SFD \cite{statefarm2020sfd} datasets.}
    \label{fig:view-consistency-crossdb}
\end{figure*}
Since the camera configuration for AUCDD-V1, EZZ2021, and SFD is very similar to that of camera D4 in the 100-Driver dataset (see Figure \ref{fig:view-consistency-crossdb}), we perform inference on these three datasets using the model trained on images acquired from cameras D1, D2, and D3. Thus, we can state that the current setup constitutes a cross-camera and cross-dataset experiment. The experimental results shown in Table \ref{tab:cross-dataset} highlight the performance of different models on the D4 test set of 100-Driver and the AUCDD-V1, EZZ2021, and SFD datasets. DBMNet achieves the highest Top-1 accuracy on the D4 set (11.29\%) and SFD (9.86\%), and ranks second on AUCDD-V1 (7.71\%) and EZZ2021 (10.51\%). MambaVision leads on AUCDD-V1 (8.64\%) and EZZ2021 (12.74\%). SwinT stands out with a strong performance on the D4 set (24.07\% Top-1), significantly higher than all other models, but underperforms on EZZ2021 (7.50\%) and AUCDD-V1 (8.17\%). MobileViT delivers a solid mid-range performance, with Top-1 scores of 9.58\% on D4 and 9.87\% on EZZ2021—higher than GhostNet and Hu \etal, but trailing behind DBMNet, SwinT, and MambaVision. In terms of Top-5 accuracy, DBMNet performs best on the SFD dataset (38.62\%) and D4 set (39.31\%), while DGCCL dominates on AUCDD-V1 (38.58\%) and EZZ2021 (34.96\%). SwinT again shines on D4 with the highest Top-5 accuracy (48.02\%) and offers solid performance on SFD (31.15\%). MobileViT performs competitively in Top-5 accuracy across AUCDD-V1 (29.28\%) and SFD (25.66\%), indicating its reasonable generalization, albeit at a higher computational cost. Si-CA MobileNet, while efficient, remains behind in both Top-1 and Top-5 accuracy compared to DBMNet, SwinT, and MambaVision. For example, on EZZ2021, it scores 10.23\% Top-1, compared to DBMNet’s 10.51\% and MambaVision’s 12.74\%. Overall, the results highlight DBMNet as a balanced and practical model for cross-camera and cross-dataset tasks—delivering strong accuracy with minimal overhead. While SwinT achieves peak accuracy in some cases, it’s far more complex. MobileViT offers a compromise, but DBMNet provides better generalization at lower cost.
\begin{table*}
    \centering
    \caption{Top-1 and Top-5 accuracy (\%) on the D4 test set of 100-Driver, and the AUCDD-V1, EZZ2021 and SFD datasets for the model trained on the combination of D1, D2 and D3 cameras of the 100-Driver dataset. In each column, the best and second best results are marked in boldface and underlined, respectively.}
    \label{tab:cross-dataset}
    \begin{tabular}{lcccc|cccc}
        \toprule
        & \multicolumn{4}{c}{Top-1} & \multicolumn{4}{c}{Top-5} \\
        & 100-Driver & AUCDD-V1 & EZZ2021 & SFD & 100-Driver & AUCDD-V1 & EZZ2021 & SFD \\ \midrule
        DGCCL \cite{yang2025domain} & 10.78 & 6.19 & \underline{11.94} & 8.77 & 34.63 & \textbf{38.58} & \textbf{34.96} & \underline{35.51} \\
        Duan \etal~\cite{duan2023enhancing} & 7.25 & 6.42 & 8.33 & 9.31 & 30.17 & 30.15 & 30.88 & 28.87 \\
        GhostNet \cite{wang2023100} & 8.85 & 6.79 & 8.08 & 4.59 & 32.35 & 29.62 & 31.80 & 23.20 \\
        Hu \etal~\cite{Hu_Driver2023} & 5.45 & 5.75 & 8.10 & 4.84 & 29.17 & 27.61 & 21.53 & 24.31 \\
        MambaVision \cite{hatamizadeh2025mambavision} & 10.81 & \textbf{8.64} & \textbf{12.74} & 6.21 & 34.75 & 29.25 & 32.37 & 26.53 \\
        MobileViT~\cite{mehtamobilevit} & 9.58 & 6.74 & 9.87 & 6.55 & 32.48 & 29.28 & 26.30 & 25.66 \\
        OLCMNet \cite{Li_Driver2022} & 8.48 & 5.56 & 9.94 & 4.96 & 30.17 & 28.09 & 27.62 & 25.11 \\ 
        Si-CA MobileNet \cite{lv2025si} & 5.35 & 5.52 & 10.23 & 8.20 & 26.78 & 28.15 & 29.18 & 29.73 \\
        SwinT~\cite{liu2021swin} & \textbf{24.07} & \underline{8.17} & 7.50 & \textbf{10.67} & \textbf{48.02} & 29.83 & 33.07 & 31.15 \\
        DBMNet (ours) & \underline{11.29} & 7.71 & 10.51 & \underline{9.86} & \underline{39.31} & \underline{30.80} & \underline{34.09} & \textbf{38.62} \\
        \bottomrule
    \end{tabular}
\end{table*}
\subsubsection{Analysis}
\begin{figure*}
    \centering
    \begin{tabular}{cc}
         \includegraphics[width=.46\linewidth]{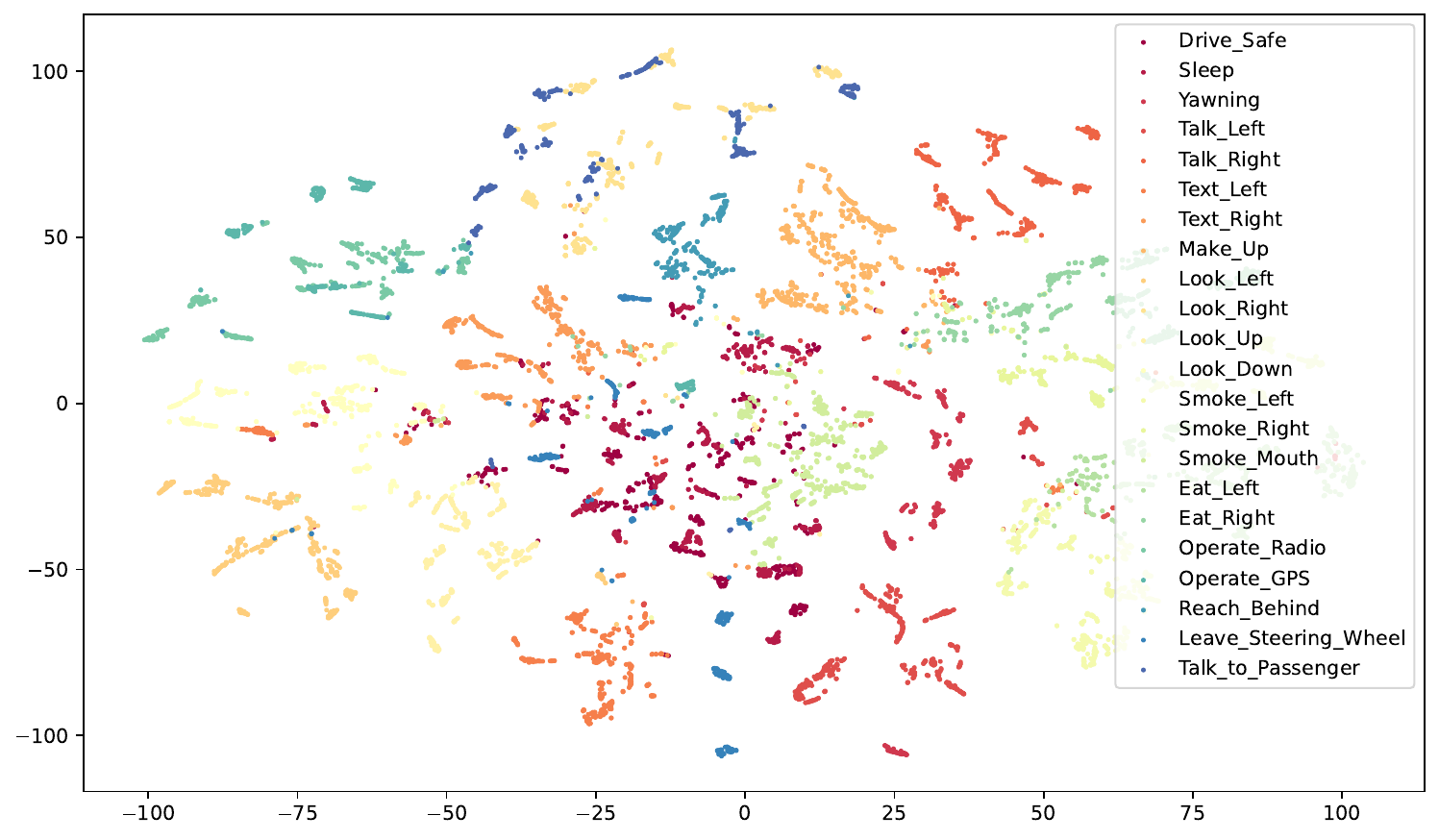} & \includegraphics[width=.46\linewidth]{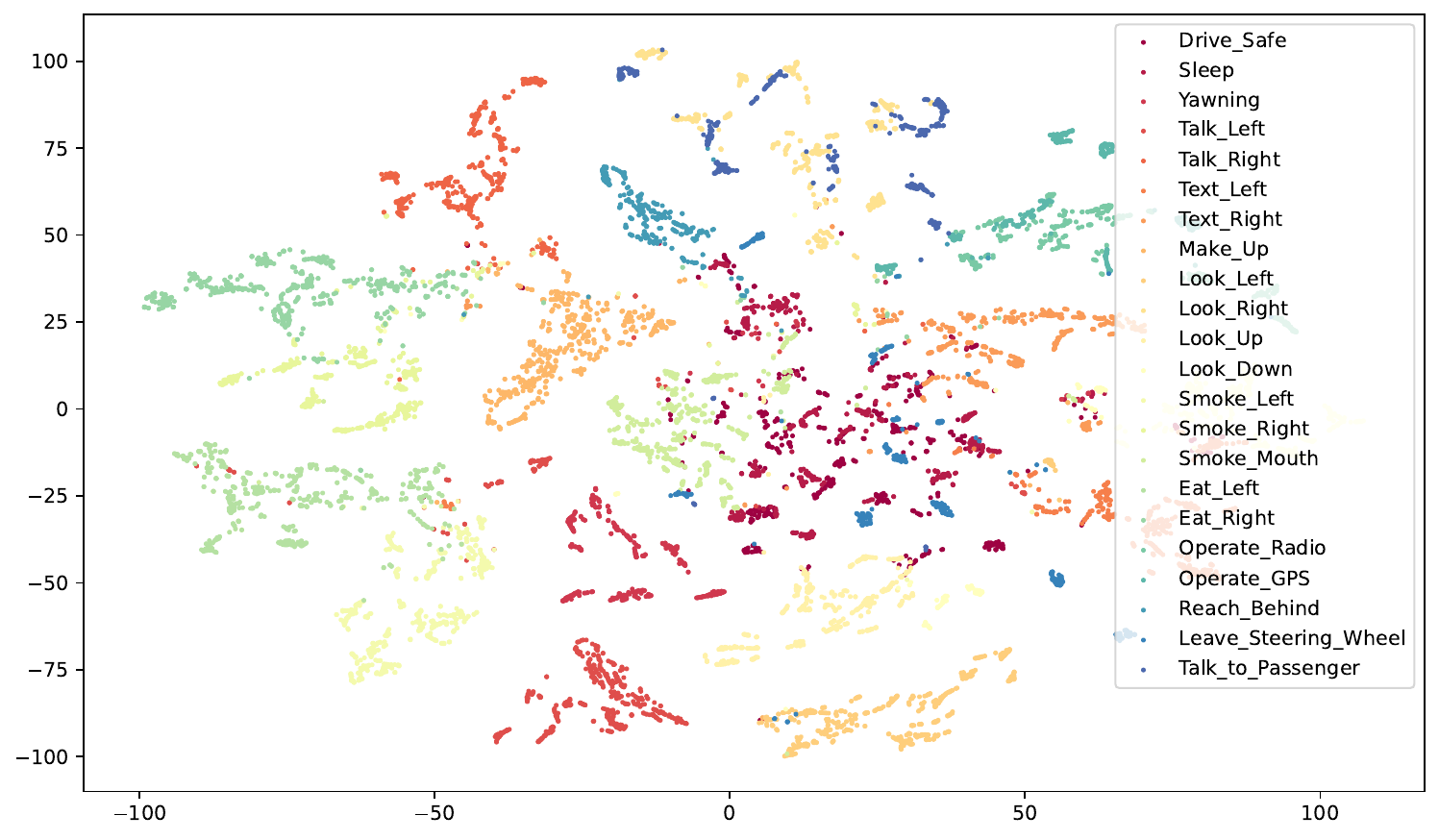} \\
         (a) & (b)
    \end{tabular}
    \caption{Analysis of features learned by our DBMNet using t-Distributed Stochastic Neighbor Embedding (t-SNE). Using the model trained on the \{D1,D2,D3\} images, we report for the validation images (a) the feature vectors $\mathbf{f}$ extracted by the backbone, and (b) the feature vectors $\mathbf{\hat{f}}$ after the disentanglement module.}
    \label{fig:tsne}
\end{figure*}

In this section, we conduct an analysis of features to support our claim of disentangled features. To this end, we qualitatively evaluate how the features before and after disentanglement module separate in the embedding space and quantitatively estimate the capacity of discriminating viewpoints before and after feature disentanglement. Practically, we extract feature vectors before and after the View disentanglement module, namely $\mathbf{f}$ and $\mathbf{\hat{f}}$, for both training and validation images. We chose validation images because the camera view categories overlap with those of the training set, unlike those of the test set. Figure \ref{fig:tsne} illustrates the features of the validation set for the image subset {D1, D2, D3} using t-Distributed Stochastic Neighbor Embedding (t-SNE). The colors represent different driver action classes. As depicted, the \(\mathbf{f}\)-features separate into small, distinct clusters, even for the same action class, likely due to the actions being captured by different cameras. In contrast, the \(\mathbf{\hat{f}}\)-features form larger clusters that more effectively distinguish between different actions regardless of the camera used, thereby demonstrating the effectiveness of the feature disentanglement module. We then exploit a Minimum Distance Classifier (MDC) to classify camera viewpoints. Specifically, we train an MDC classifier on $\mathbf{f}$ feature vectors, referred as $\mathrm{MDC}_\mathbf{f}$, and another MDC classifier on $\mathbf{\hat{f}}$ feature vectors, labeled as $\mathrm{MDC}_\mathbf{\hat{f}}$. Finally, we classify $\mathbf{f}$ validation samples with $\mathrm{MDC}_\mathbf{f}$ and $\mathbf{\hat{f}}$ validation samples with $\mathrm{MDC}_\mathbf{\hat{f}}$. Results are reported in Table \ref{tab:analysis}.
\begin{table}
    \centering
    \caption{Results of an MDC on the validation set for the different cross-camera setups.}
    \label{tab:analysis}
    \begin{tabular}{l|c}
    \toprule
                                       & View cls acc. ($\mathbf{f} \rightarrow \mathbf{\hat{f}}$) \\ \midrule
         \{D2,D3,D4\} $\rightarrow$ D1 &  31.61 $\rightarrow$ 19.52 \\
         \{D1,D3,D4\} $\rightarrow$ D2 &  32.06 $\rightarrow$ 20.95 \\
         \{D1,D2,D4\} $\rightarrow$ D3 & 44.86 $\rightarrow$ 19.87 \\
         \{D1,D2,D3\} $\rightarrow$ D4 & 70.83 $\rightarrow$ 34.75 \\ \bottomrule
    \end{tabular}
\end{table}
As can be seen, the accuracy of view classification undergoes a drop after feature disentanglement. On average, there is a decrease of 21.07\%, with the most significant drop for D4 at 34.08\% and the least pronounced for Camera 1 and 2, around 11.05\%. This means that view information is discarded in favor of a better ability to discriminate drivers' actions.

\subsubsection{Discussion}
\begin{figure}
    \centering
    \includegraphics[width=\linewidth]{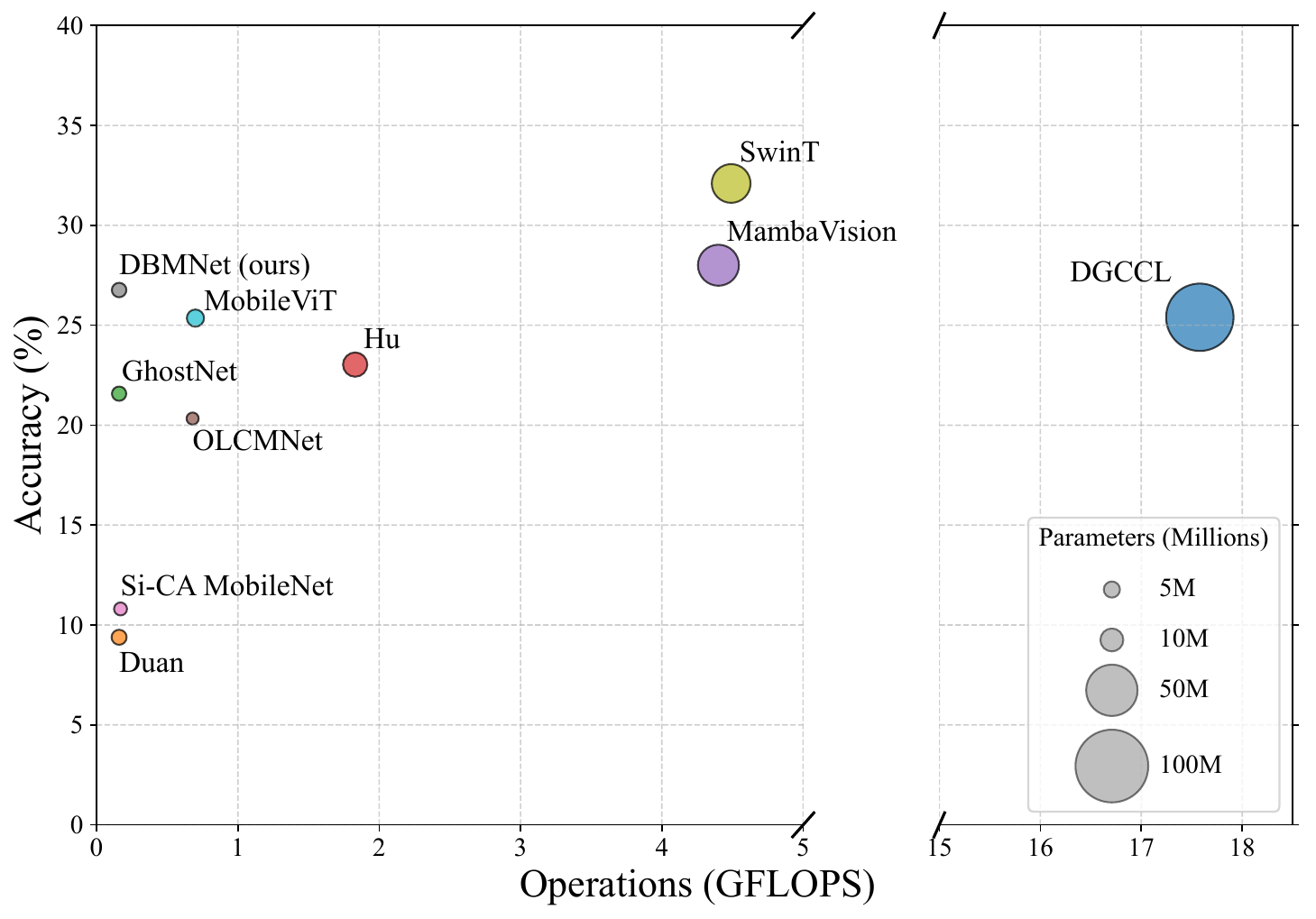}
    \caption{Ball chart illustrating the trade-off between computational complexity (measured in GFLOPS) and average accuracy obtained by the compared methods across all cross-camera and cross-dataset experiments. The size of each ball is proportional to the number of model parameters.}
    \label{fig:cumulative-acc}
\end{figure}

In this section, we summarize the results of both the cross-dataset and cross-camera experiments, incorporating an analysis from a computational efficiency perspective. The ball plot in Figure~\ref{fig:cumulative-acc} illustrates the trade-off between accuracy, computational cost (GFLOPS), and model size (bubble diameter). Based on the ball chart, the overall results highlight a clear trade-off between model complexity and recognition accuracy. It is important to note that the accuracy reported is the average accuracy across all cross-dataset and cross-camera experiments. Lightweight models like GhostNet, Duan \etal, and Si-CA MobileNet have minimal computational costs ($\le$0.17 \text{GFLOPs}) and small model sizes ($\le$4.3M parameters), but they suffer from lower accuracy—ranging from 9.39\% (Duan) to 21.58\% (GhostNet). On the other extreme, DGCCL and MambaVision achieve relatively high accuracy (25.40\% and 28.00\%, respectively) but at the cost of massive complexity: DGCCL tops out at 85.8M parameters and 17.58 GFLOPs, while MambaVision requires 31.8M parameters and 4.40 GFLOPs. SwinT and MobileViT also perform well accuracy-wise, with SwinT reaching the highest average (29.55\%) but demanding 28.3M parameters and 4.49 GFLOPs. MobileViT achieves 23.69\% accuracy at a more moderate 5.6M parameters and 0.70 GFLOPs. DBMNet stands out by achieving high accuracy (26.76\%) with minimal computational cost (0.16 GFLOPs) and a compact size (4.0M parameters). This places it near the top in performance while staying among the most efficient models—offering a compelling balance of speed, size, and predictive power, well-suited for real-world, resource-constrained applications.


\subsection{Deployment on embedded device}
The proposed DBMNet and the competing methods for distracted driver classification are deployed on the Coral Dev Board 4GB, a commercially available ARM-based Single-Board Computer (SBC) designed specifically for edge AI applications~\cite{coral-board}. This development board integrates a Google Edge TPU coprocessor, a specialized hardware accelerator optimized for efficient execution of deep learning models directly on the device.

The Edge TPU is an 8-bit fixed-point coprocessor capable of performing high-speed inference on quantized TensorFlow Lite models with low power consumption. This makes it particularly suitable for real-time, on-device inference tasks, such as monitoring driver distraction. The Coral Dev Board therefore enables compact, energy-efficient, and responsive deployment of the proposed system in embedded automotive environments.

All methods are deployed following post-training quantization using full integer quantization. This optimization technique converts both the model weights and activations from 32-bit floating-point to 8-bit fixed-point representations. It requires calibration with a representative dataset to determine optimal quantization parameters for the activations. For this purpose, we use the validation set of the 100-Driver dataset in a leave-one-camera-out configuration. A quantized version of each model is generated for every configuration of daylight data. Due to the limited computational resources of the Coral Dev Board, we were unable to successfully deploy the DGCCL, MambaVision, and SwinT models. Therefore, these models have been excluded from this evaluation.

Performance is evaluated across five metrics:
\begin{itemize}
    \item Average error represents the classification error obtained across the four folds of the leave-one-camera-out experiment;
    \item Model size refers to the size of the quantized model file (including architecture and weights), reported in megabytes (MB);
    \item Memory usage is the average memory consumption (in MB) during inference, computed over 100 inference passes;
    \item Inference time is the average time (in seconds) required for a single forward pass, again averaged over 100 runs;
    \item Power consumption is estimated by measuring the energy used across 100 inference passes, scaled by the average inference time, and reported in milliampere-hours (mAh).
\end{itemize}

\begin{figure}
    \centering
    \includegraphics[width=\columnwidth]{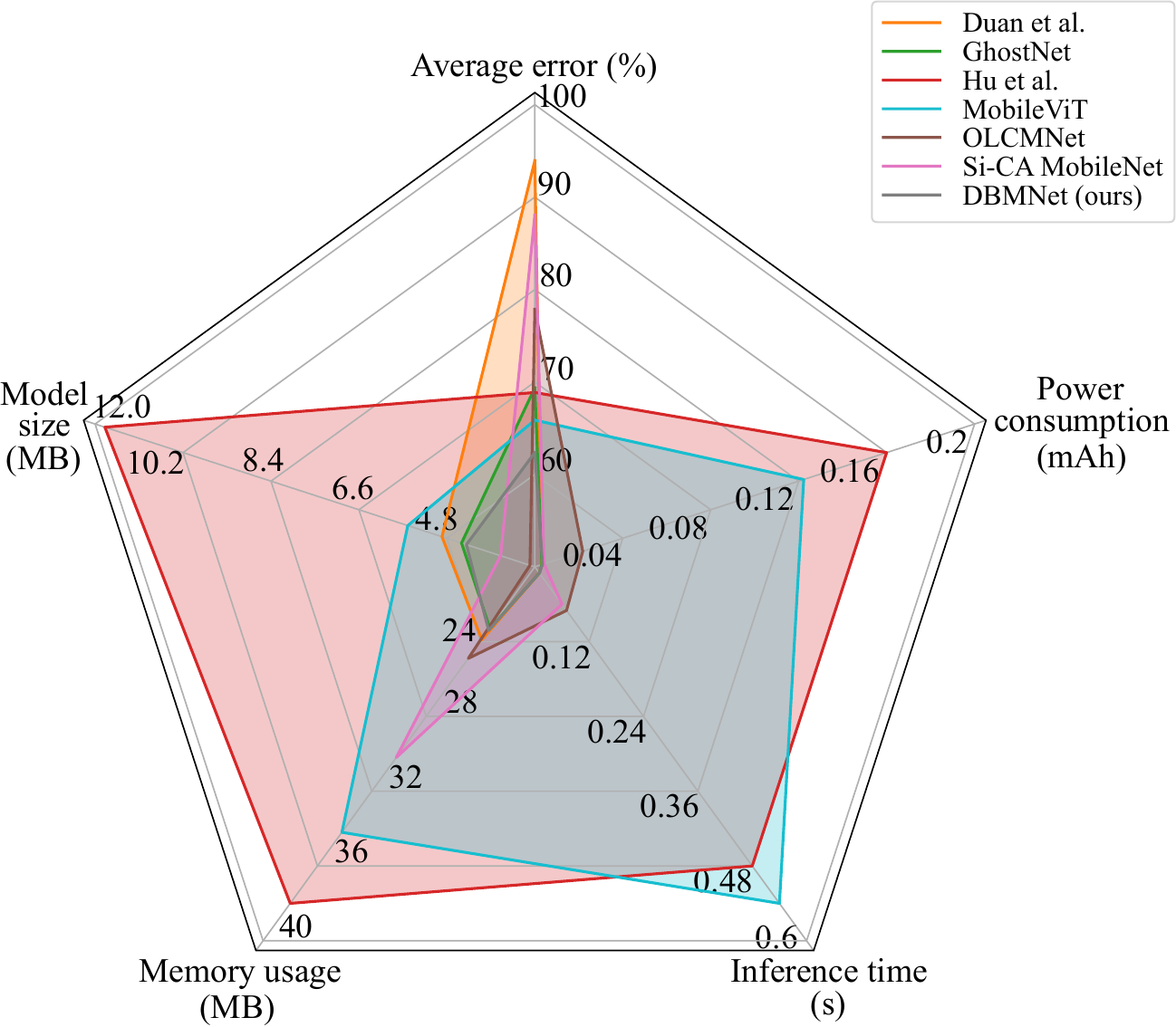}
    \caption{Radar plot comparing the proposed DBMNet with competing methods deployed on the Coral Dev Board across five key metrics: average error, model size, memory usage, inference time, and power consumption.}
    \label{fig:radar-plot}
\end{figure}

The radar plot reported in Figure \ref{fig:radar-plot} highlights DBMNet (ours) as the best-performing model, achieving the lowest average error while maintaining a compact model size, low memory usage, fast inference time, and minimal power consumption -- making it highly suitable for real-time, embedded applications. Compared to other methods, DBMNet offers the most balanced trade-off between accuracy and efficiency. Duan \etal, GhostNet and Si-CA MobileNet are similarly lightweight and power-efficient but suffer from significantly higher error rates. Hu \etal~is the most resource-intensive model in terms of size, memory, and power, without delivering a proportional gain in accuracy. OLCMNet performs moderately across all metrics but doesn't match DBMNet's overall efficiency-accuracy balance. MobileViT, while offering better accuracy than most lightweight models, shows higher memory usage, slower inference, and greater power consumption—making it less suitable for constrained environments despite its solid predictive performance.


\section{Conclusions}
Driver Monitoring Systems (DMSs) need to be robust to changes in vehicle configuration and camera positions. In this paper, we introduced DBMNet, a novel network designed for distracted driver classification, ensuring consistent performance regardless of camera placement. Given that DMSs operate on embedded devices with constrained computational resources, our model exploits GhostNet, a lightweight CNN, as the backbone. We improved the invariance to the camera view of the representation using a feature disentanglement module. In addition, we employed a supervised contrastive learning approach to increase consistency between intra-class representations and differentiate inter-class representations.

Extensive experiments on the 100-Driver dataset demonstrated that DBMNet achieves a 9\% improvement in Top-1 accuracy over state-of-the-art methods. Furthermore, cross-dataset evaluations on three benchmark datasets -- AUCDD-V1, EZZ2021 and SFD -- confirmed the strong generalization capabilities of our approach. Overall, the proposed method achieves a 7\% gain in Top-1 accuracy compared to existing efficient approaches. Through the analysis of features both before and after the feature disentanglement module, we determined that the objective of discarding view information is accomplished.

To validate practical deployability, we implemented DBMNet and competing models on the Coral Dev Board. In this embedded setting, DBMNet exhibited the lowest average error, along with low memory usage, fast inference time, compact model size, and minimal power consumption. These advantages are particularly relevant when contrasted with large-scale models such as SwinT and MambaVision. While such models can reach higher accuracy under ideal conditions, their heavy compuadtational and memory requirements make them impractical for embedded automotive systems. DBMNet, by contrast, strikes a better balance between performance and efficiency—meeting the strict real-time and power constraints of in-vehicle deployment.

Despite these strengths, some limitations remain. Due to the complexity of the task, overall accuracy is still modest, especially for ``Normal driving'' class, which is frequently misclassified as a distraction category. Additionally, the inference time of 1.15 × $10^{-2}$ seconds on board is relatively high for real-time safety-critical applications.

To address these issues, future work will focus on exploring lighter backbones beyond GhostNet to reduce inference time without sacrificing performance. We also plan to investigate few-shot learning strategies for model calibration on unseen camera viewpoints using limited labeled data. These improvements aim to enhance both the efficiency and adaptability of DBMNet in real-world DMS deployments.

Overall, our work highlights the critical importance of cross-camera generalization for real-world driver monitoring systems and provides a novel solution through a lightweight, disentanglement-based model rigorously evaluated under a realistic leave-one-camera-out protocol. This contributes a new perspective to the field by explicitly addressing viewpoint variability—an underexplored but practically crucial challenge.

\section*{Acknowledgment}

Financial support from ICSC – Centro Nazionale di Ricerca in High Performance Computing, Big Data and Quantum Computing, funded by European Union – NextGenerationEU.

\ifCLASSOPTIONcaptionsoff
  \newpage
\fi



\bibliographystyle{IEEEtran}

\begin{thebibliography}{10}
\providecommand{\url}[1]{#1}
\csname url@samestyle\endcsname
\providecommand{\newblock}{\relax}
\providecommand{\bibinfo}[2]{#2}
\providecommand{\BIBentrySTDinterwordspacing}{\spaceskip=0pt\relax}
\providecommand{\BIBentryALTinterwordstretchfactor}{4}
\providecommand{\BIBentryALTinterwordspacing}{\spaceskip=\fontdimen2\font plus
\BIBentryALTinterwordstretchfactor\fontdimen3\font minus
  \fontdimen4\font\relax}
\providecommand{\BIBforeignlanguage}[2]{{%
\expandafter\ifx\csname l@#1\endcsname\relax
\typeout{** WARNING: IEEEtran.bst: No hyphenation pattern has been}%
\typeout{** loaded for the language `#1'. Using the pattern for}%
\typeout{** the default language instead.}%
\else
\language=\csname l@#1\endcsname
\fi
#2}}
\providecommand{\BIBdecl}{\relax}
\BIBdecl

\bibitem{world2023global}
{World Health Organization}, ``Global status report on road safety 2023,''
  \emph{World Health Organization}, 2023.

\bibitem{road-mortality}
{The Global Health Observatory}, ``Road traffic mortality,''
  \url{https://www.who.int/data/gho/data/themes/topics/topic-details/GHO/road-traffic-mortality},
  2023, (accessed: Jun 18, 2025).

\bibitem{driver-distraction-summary}
{European Road Safety Observatory}, ``Driver distraction summary,''
  \url{https://road-safety.transport.ec.europa.eu/system/files/2021-07/ersosynthesis2018-driverdistraction-summary.pdf},
  2018, (accessed: Jun 18, 2025).

\bibitem{choi2016driver}
Y.~Choi, S.~I. Han, S.-H. Kong, and H.~Ko, ``Driver status monitoring systems
  for smart vehicles using physiological sensors: A safety enhancement system
  from automobile manufacturers,'' \emph{IEEE Signal Processing Magazine},
  vol.~33, no.~6, pp. 22--34, 2016.

\bibitem{eu2019}
{European Parliment and Council}, ``Regulation (eu) 2019/2144 of the european
  parliament and of the council,''
  \url{https://eur-lex.europa.eu/eli/reg/2019/2144/oj}, 2019, on type-approval
  requirements for motor vehicles with regard to their general safety and the
  protection of vehicle occupants and vulnerable road users (accessed: Jun 18,
  2025).

\bibitem{eu2023}
------, ``Commission delegated regulation (eu) 2023/2590,''
  \url{https://eur-lex.europa.eu/eli/reg_del/2023/2590/oj}, 2023, supplementing
  Regulation (EU) 2019/2144 with technical requirements for advanced driver
  distraction warning systems (accessed: Jun 18, 2025).

\bibitem{eraqi2019driver}
H.~M. Eraqi, Y.~Abouelnaga, M.~H. Saad, M.~N. Moustafa \emph{et~al.}, ``Driver
  distraction identification with an ensemble of convolutional neural
  networks,'' \emph{Journal of Advanced Transportation}, vol. 2019, 2019.

\bibitem{wang2023100}
J.~Wang, W.~Li, F.~Li, J.~Zhang, Z.~Wu, Z.~Zhong, and N.~Sebe, ``100-driver: A
  large-scale, diverse dataset for distracted driver classification,''
  \emph{IEEE Trans. Intell. Transport. Syst.}, 2023.

\bibitem{Hu_Driver2023}
Z.~Hu, Y.~Xing, W.~Gu, D.~Cao, and C.~Lv, ``Driver anomaly quantification for
  intelligent vehicles: A contrastive learning approach with representation
  clustering,'' \emph{IEEE Trans. Intell. Veh.}, vol.~8, no.~1, pp. 37--47,
  2023.

\bibitem{bianco2023platform}
S.~Bianco, L.~Celona, G.~D. Gallo, and P.~Napoletano, ``A platform for
  multi-modal driver behaviour data collection,'' in \emph{International
  Conference on Consumer Electronics-Berlin}.\hskip 1em plus 0.5em minus
  0.4em\relax IEEE, 2023, pp. 112--116.

\bibitem{beheraLatentBodyPoseGuided2018}
A.~Behera and A.~H. Keidel, ``Latent {{Body-Pose}} guided {{DenseNet}} for
  {{Recognizing Driver}}'s {{Fine-grained Secondary Activities}},'' in
  \emph{IEEE Int. Conf. Adv. Video Signal-Based Surveill. (AVSS)}.\hskip 1em
  plus 0.5em minus 0.4em\relax {IEEE}, 2018, pp. 1--6.

\bibitem{ezzouhri2021robust}
A.~Ezzouhri, Z.~Charouh, M.~Ghogho, and Z.~Guennoun, ``Robust deep
  learning-based driver distraction detection and classification,'' \emph{IEEE
  Access}, vol.~9, pp. 168\,080--168\,092, 2021.

\bibitem{statefarm2020sfd}
\BIBentryALTinterwordspacing
``Distracted dirver detection competition,'' 2020. [Online]. Available:
  \url{https://www.kaggle.com/c/state-farm-distracted-driver-detection}
\BIBentrySTDinterwordspacing

\bibitem{Abouelnaga_Realtime2018}
Y.~Abouelnaga, H.~M. Eraqi, and M.~N. Moustafa, ``Real-time distracted driver
  posture classification,'' in \emph{Proc. Conf. Neural Inf. Process. Syst.
  (NIPS)}, 2018, pp. 1--8.

\bibitem{Baheti_Detection2018}
B.~Baheti, S.~Gajre, and S.~Talbar, ``Detection of distracted driver using
  convolutional neural network,'' in \emph{CVPRW}.\hskip 1em plus 0.5em minus
  0.4em\relax IEEE, 2018, pp. 1145--1151.

\bibitem{Koay_Convolutional2021}
H.~V. Koay, J.~H. Chuah, and C.-O. Chow, ``Convolutional neural network or
  vision transformer? benchmarking various machine learning models for
  distracted driver detection,'' in \emph{IEEE Reg. 10 Annu. Int. Conf. Proc.
  TENCON}.\hskip 1em plus 0.5em minus 0.4em\relax IEEE, 2021, pp. 417--422.

\bibitem{Shaout_Embedded2021}
A.~Shaout, B.~Roytburd, and L.~A. {Sanchez-Perez}, ``An embedded deep learning
  computer vision method for driver distraction detection,'' in \emph{Int. Arab
  Conf. Inf. Technol., (ACIT)}.\hskip 1em plus 0.5em minus 0.4em\relax IEEE,
  2021, pp. 1--7.

\bibitem{Baheti_Computationally2020}
B.~Baheti, S.~Talbar, and S.~Gajre, ``Towards computationally efficient and
  realtime distracted driver detection with mobilevgg network,'' \emph{IEEE
  Trans. Intell. Veh.}, vol.~5, no.~4, pp. 565--574, 2020.

\bibitem{Duy-LinhNguyen_Driver2022}
{Duy-Linh Nguyen}, M.~D. Putro, and K.-H. Jo, ``Driver behaviors recognizer
  based on light-weight convolutional neural network architecture and attention
  mechanism,'' \emph{IEEE Access}, vol.~10, pp. 71\,019--71\,029, 2022.

\bibitem{Li_Driver2022}
P.~Li, Y.~Yang, R.~Grosu, G.~Wang, R.~Li, Y.~Wu, and Z.~Huang, ``Driver
  distraction detection using octave-like convolutional neural network,''
  \emph{IEEE Trans. Intell. Transport. Syst.}, vol.~23, no.~7, pp. 8823--8833,
  2022.

\bibitem{Liu_Extremely2023}
D.~Liu, T.~Yamasaki, Y.~Wang, K.~Mase, and J.~Kato, ``Toward extremely
  lightweight distracted driver recognition with distillation-based neural
  architecture search and knowledge transfer,'' \emph{IEEE Trans. Intell.
  Transport. Syst.}, vol.~24, no.~1, pp. 764--777, 2023.

\bibitem{Mittal_CATCapsNet2023}
H.~Mittal and B.~Verma, ``{CAT-CapsNet}: A convolutional and attention based
  capsule network to detect the driver's distraction,'' \emph{IEEE Trans.
  Intell. Transport. Syst.}, vol.~24, no.~9, pp. 9561--9570, 2023.

\bibitem{leekhaAreYouPaying2019}
M.~Leekha, M.~Goswami, R.~R. Shah, Y.~Yin, and R.~Zimmermann, ``Are {{You
  Paying Attention}}? {{Detecting Distracted Driving}} in {{Real-Time}},'' in
  \emph{International Conference on Multimedia Big Data}.\hskip 1em plus 0.5em
  minus 0.4em\relax {IEEE}, 2019, pp. 171--180.

\bibitem{Xing_Driver2019}
Y.~Xing, C.~Lv, H.~Wang, D.~Cao, E.~Velenis, and F.-Y. Wang, ``Driver activity
  recognition for intelligent vehicles: A deep learning approach,'' \emph{IEEE
  Trans. Veh. Technol.}, vol.~68, no.~6, pp. 5379--5390, Jun. 2019.

\bibitem{Qin_Distracted2022}
B.~Qin, J.~Qian, Y.~Xin, B.~Liu, and Y.~Dong, ``Distracted driver detection
  based on a cnn with decreasing filter size,'' \emph{IEEE Trans. Intell.
  Transport. Syst.}, vol.~23, no.~7, pp. 6922--6933, 2022.

\bibitem{deyContextdrivenDetectionDistracted2021}
A.~K. Dey, B.~Goel, and S.~Chellappan, ``Context-driven detection of distracted
  driving using images from in-car cameras,'' \emph{Internet Things}, vol.~14,
  2021.

\bibitem{Behera_Deep2022}
A.~Behera, Z.~Wharton, A.~Keidel, and B.~Debnath, ``Deep cnn, body pose, and
  body-object interaction features for drivers' activity monitoring,''
  \emph{IEEE Trans. Intell. Transport. Syst.}, vol.~23, no.~3, pp. 2874--2881,
  2022.

\bibitem{wangDataAugmentationApproach2021}
J.~Wang, Z.~Wu, F.~Li, and J.~Zhang, ``A {{Data Augmentation Approach}} to
  {{Distracted Driving Detection}},'' \emph{Future Internet}, vol.~13, no.~1,
  2021.

\bibitem{liNovelSpatialTemporalGraph2019}
P.~Li, M.~Lu, Z.~Zhang, D.~Shan, and Y.~Yang, ``A {{Novel Spatial-Temporal
  Graph}} for {{Skeleton-based Driver Action Recognition}},'' in
  \emph{Intelligent Transportation Systems Conference}.\hskip 1em plus 0.5em
  minus 0.4em\relax {IEEE}, 2019, pp. 3243--3248.

\bibitem{Bera_Attend2021}
A.~Bera, Z.~Wharton, Y.~Liu, N.~Bessis, and A.~Behera, ``{Attend and Guide
  (AG-Net)}: A keypoints-driven attention-based deep network for image
  recognition,'' \emph{IEEE TIP}, vol.~30, pp. 3691--3704, 2021.

\bibitem{duan2023enhancing}
C.~Duan, Z.~Liu, J.~Xia, M.~Zhang, J.~Liao, and L.~Cao, ``Enhancing
  cross-dataset performance of distracted driving detection with score softmax
  classifier and dynamic gaussian smoothing supervision,'' \emph{IEEE
  Transactions on Intelligent Vehicles}, pp. 1--14, 2024.

\bibitem{Masood_Detecting2020}
S.~Masood, A.~Rai, A.~Aggarwal, M.~Doja, and M.~Ahmad, ``Detecting distraction
  of drivers using convolutional neural network,'' \emph{Pattern Recognit.
  Lett.}, vol. 139, pp. 79--85, 2020.

\bibitem{Li_Learning2022}
W.~Li, J.~Wang, T.~Ren, F.~Li, J.~Zhang, and Z.~Wu, ``Learning accurate,
  speedy, lightweight cnns via instance-specific multi-teacher knowledge
  distillation for distracted driver posture identification,'' \emph{IEEE
  Trans. Intell. Transport. Syst.}, vol.~23, no.~10, pp. 17\,922--17\,935,
  2022.

\bibitem{Peng_TransDARC2022}
K.~Peng, A.~Roitberg, K.~Yang, J.~Zhang, and R.~Stiefelhagen, ``{TransDARC}:
  Transformer-based driver activity recognition with latent space feature
  calibration,'' in \emph{IEEE Int. Conf. Intell. Rob. Syst. (IROS)}.\hskip 1em
  plus 0.5em minus 0.4em\relax IEEE, 2022, pp. 278--285.

\bibitem{mohammed2024driver}
A.~A. Mohammed, X.~Geng, J.~Wang, and Z.~Ali, ``Driver distraction detection
  using semi-supervised lightweight vision transformer,'' \emph{Engineering
  Applications of Artificial Intelligence}, vol. 129, p. 107618, 2024.

\bibitem{yang2025domain}
L.~Yang, H.~Wei, Z.~Hu, and C.~Lv, ``A domain generalization method for
  deploying driver distraction detection models to practical application
  scenarios,'' \emph{Elsevier Engineering Applications of Artificial
  Intelligence}, vol. 152, p. 110844, 2025.

\bibitem{Li_New2022}
B.~Li, J.~Chen, Z.~Huang, H.~Wang, J.~Lv, J.~Xi, J.~Zhang, and Z.~Wu, ``A new
  unsupervised deep learning algorithm for fine-grained detection of driver
  distraction,'' \emph{IEEE Trans. Intell. Transport. Syst.}, vol.~23, no.~10,
  pp. 19\,272--19\,284, 2022.

\bibitem{kopuklu2021driver}
O.~Kopuklu, J.~Zheng, H.~Xu, and G.~Rigoll, ``Driver anomaly detection: A
  dataset and contrastive learning approach,'' in \emph{Proceedings of the
  IEEE/CVF Winter Conference on Applications of Computer Vision}, 2021, pp.
  91--100.

\bibitem{koay2023contrastive}
H.~V. Koay, J.~H. Chuah, and C.-O. Chow, ``Contrastive learning with video
  transformer for driver distraction detection through multiview and multimodal
  video,'' in \emph{2023 IEEE Region 10 Symposium (TENSYMP)}.\hskip 1em plus
  0.5em minus 0.4em\relax IEEE, 2023, pp. 1--6.

\bibitem{Yang_Quantitative2023}
H.~Yang, H.~Liu, Z.~Hu, A.-T. Nguyen, T.-M. Guerra, and C.~Lv, ``Quantitative
  identification of driver distraction: A weakly supervised contrastive
  learning approach,'' \emph{IEEE Trans. Intell. Transport. Syst.}, pp. 1--12,
  2023.

\bibitem{khan2022supervised}
S.~S. Khan, Z.~Shen, H.~Sun, A.~Patel, and A.~Abedi, ``Supervised contrastive
  learning for detecting anomalous driving behaviours from multimodal videos,''
  in \emph{2022 19th Conference on Robots and Vision (CRV)}.\hskip 1em plus
  0.5em minus 0.4em\relax IEEE, 2022, pp. 16--23.

\bibitem{okon2017detecting}
O.~D. Okon and L.~Meng, ``Detecting distracted driving with deep learning,'' in
  \emph{International Conference on Interactive Collaborative Robotics}.\hskip
  1em plus 0.5em minus 0.4em\relax Springer, 2017, pp. 170--179.

\bibitem{liu2021tml}
D.~Liu, T.~Yamasaki, Y.~Wang, K.~Mase, and J.~Kato, ``Tml: A triple-wise
  multi-task learning framework for distracted driver recognition,'' \emph{IEEE
  Access}, vol.~9, pp. 125\,955--125\,969, 2021.

\bibitem{han2020ghostnet}
K.~Han, Y.~Wang, Q.~Tian, J.~Guo, C.~Xu, and C.~Xu, ``Ghostnet: More features
  from cheap operations,'' in \emph{CVPR}.\hskip 1em plus 0.5em minus
  0.4em\relax IEEE/CVF, 2020, pp. 1580--1589.

\bibitem{ha2017hypernetworks}
D.~Ha, A.~M. Dai, and Q.~V. Le, ``Hypernetworks,'' in \emph{International
  Conference on Learning Representations}, 2017.

\bibitem{chauhan2024brief}
V.~K. Chauhan, J.~Zhou, P.~Lu, S.~Molaei, and D.~A. Clifton, ``A brief review
  of hypernetworks in deep learning,'' \emph{Artificial Intelligence Review},
  vol.~57, no.~9, p. 250, 2024.

\bibitem{schug2024attention}
S.~Schug, S.~Kobayashi, Y.~Akram, J.~Sacramento, and R.~Pascanu, ``Attention as
  a hypernetwork,'' \emph{arXiv preprint arXiv:2406.05816}, 2024.

\bibitem{siddiqui2024dvanet}
N.~Siddiqui, P.~Tirupattur, and M.~Shah, ``Dvanet: Disentangling view and
  action features for multi-view action recognition,'' in \emph{Conference on
  Artificial Intelligence}, vol.~38, no.~5.\hskip 1em plus 0.5em minus
  0.4em\relax AAAI, 2024, pp. 4873--4881.

\bibitem{schroff2015facenet}
F.~Schroff, D.~Kalenichenko, and J.~Philbin, ``Facenet: A unified embedding for
  face recognition and clustering,'' in \emph{Conference on Computer Vision and
  Pattern Recognition}.\hskip 1em plus 0.5em minus 0.4em\relax IEEE, 2015, pp.
  815--823.

\bibitem{howard2019searching}
A.~Howard, M.~Sandler, G.~Chu, L.-C. Chen, B.~Chen, M.~Tan, W.~Wang, Y.~Zhu,
  R.~Pang, V.~Vasudevan \emph{et~al.}, ``Searching for mobilenetv3,'' in
  \emph{ICCV}.\hskip 1em plus 0.5em minus 0.4em\relax IEEE/CVF, 2019, pp.
  1314--1324.

\bibitem{iandola2016squeezenet}
F.~N. Iandola, S.~Han, M.~W. Moskewicz, K.~Ashraf, W.~J. Dally, and K.~Keutzer,
  ``Squeezenet: Alexnet-level accuracy with 50x fewer parameters and< 0.5 mb
  model size,'' \emph{preprint arXiv:1602.07360}, 2016.

\bibitem{liu2024image}
Y.~Liu, J.~Xue, D.~Li, W.~Zhang, T.~K. Chiew, and Z.~Xu, ``Image recognition
  based on lightweight convolutional neural network: Recent advances,''
  \emph{Image and Vision Computing}, p. 105037, 2024.

\bibitem{zhong2020random}
Z.~Zhong, L.~Zheng, G.~Kang, S.~Li, and Y.~Yang, ``Random erasing data
  augmentation,'' in \emph{AAAI Conference on Artificial Intelligence},
  vol.~34, 2020, pp. 13\,001--13\,008.

\bibitem{hatamizadeh2025mambavision}
A.~Hatamizadeh and J.~Kautz, ``Mambavision: A hybrid mamba-transformer vision
  backbone,'' in \emph{Computer Vision and Pattern Recognition
  Conference}.\hskip 1em plus 0.5em minus 0.4em\relax IEEE/CVF, 2025, pp.
  25\,261--25\,270.

\bibitem{mehtamobilevit}
S.~Mehta and M.~Rastegari, ``Mobilevit: Light-weight, general-purpose, and
  mobile-friendly vision transformer,'' in \emph{International Conference on
  Learning Representations}, 2022.

\bibitem{lv2025si}
M.~Lv, Y.~Liu, Z.~Zha, X.~Zheng, H.~Wang, Y.~Wen, and Z.~Guo, ``Si-ca
  mobilenet: A lightweight and efficient convolutional neural network for
  distracted driver detection,'' \emph{Elsevier Neurocomputing}, p. 131281,
  2025.

\bibitem{liu2021swin}
Z.~Liu, Y.~Lin, Y.~Cao, H.~Hu, Y.~Wei, Z.~Zhang, S.~Lin, and B.~Guo, ``Swin
  transformer: Hierarchical vision transformer using shifted windows,'' in
  \emph{International Conference on Computer Vision}.\hskip 1em plus 0.5em
  minus 0.4em\relax IEEE/CVF, 2021, pp. 10\,012--10\,022.

\bibitem{coral-board}
Google, ``Dev board datasheet,''
  \url{https://coral.ai/docs/dev-board/datasheet}, 2020, (accessed: Jun 18,
  2025).

\end{thebibliography}

%

%

\begin{IEEEbiography}[{\includegraphics[width=1in,height=1.25in,clip,keepaspectratio]{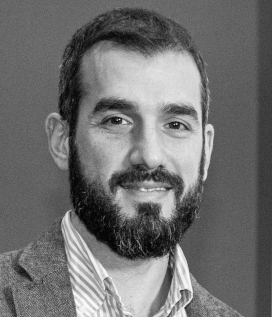}}]{Luigi Celona} is an Assistant Professor of computer science at the University of Milano-Bicocca, Italy. In 2018 and 2014, he obtained respectively the PhD and the MSc degree in Computer Science at DISCo. In 2011, he obtained the BSc degree in Computer Science from the University of Messina. His current research interests focus on image analysis and classification, ML and face analysis. He is a member of the European Laboratory for Learning and Intelligent Systems (ELLIS Society). He is the Secretary of the IEEE CTSoc Machine Learning, Deep Learning and AI in CE (MDA) Technical Committee (TC).
\end{IEEEbiography}

\begin{IEEEbiography}[{\includegraphics[width=1in,height=1.25in,clip,keepaspectratio]{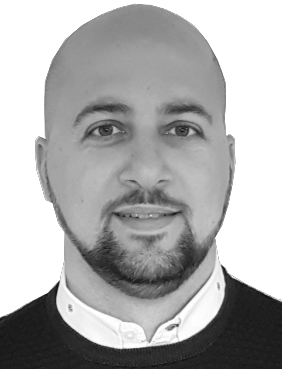}}]{Simone Bianco} is currently an Associate Professor of computer science with the University of Milano-Bicocca, Milan, Italy. He is on Stanford University's World Ranking Scientists List for his achievements in Artificial Intelligence (AI) and Image Processing. His teaching and research interests include computer vision, AI, Machine Learning (ML), optimization algorithms applied in multimodal, and multimedia applications. He is also a R\&D Manager of the University of Milano-Bicocca spin-off Imaging and Vision Solutions, and Member of European Laboratory for Learning and Intelligent Systems (ELLIS Society).
\end{IEEEbiography}

\begin{IEEEbiography}[{\includegraphics[width=1in,height=1.25in,clip,keepaspectratio]{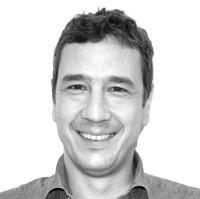}}]{Paolo Napoletano} (Member, IEEE) is an Associate Professor of computer science at the University of Milano-Bicocca, since 2021. His main research interests include AI, ML and DL, CV, PR, intelligent sensors, biological signal processing, and human–machine systems. He is a member of the European Laboratory for Learning and Intelligent Systems (ELLIS Society). He is an Associate Editor of IEEE Journal of Biomedical and Health Informatics, Neurocomputing (Elsevier), IET Signal Processing, Sensors (MDPI), and Smart Cities (MDPI). He is the Chair of the IEEE CTSoc Machine Learning, Deep Learning and AI in CE (MDA) Technical Committee (TC).

\end{IEEEbiography}







\end{document}